\documentclass[sigconf]{acmart}

\AtBeginDocument{%
  }

\usepackage{algorithm}
\usepackage{algorithmicx}
\usepackage{algpseudocode} 
\usepackage{bm}
\usepackage{multirow}
\usepackage{subfigure}
\usepackage{enumitem}
\usepackage{booktabs}
\usepackage{xcolor}

\usepackage[autostyle]{csquotes}
\newtheorem{theorem}{Theorem}

\copyrightyear{2025}
\acmYear{2025}
\setcopyright{cc}
\setcctype{by}
\acmConference[ICMR '25]{Proceedings of the 2025 International Conference on Multimedia Retrieval}{June 30-July 3, 2025}{Chicago, IL, USA}
\acmBooktitle{Proceedings of the 2025 International Conference on Multimedia Retrieval (ICMR '25), June 30-July 3, 2025, Chicago, IL, USA}
\acmDOI{10.1145/3731715.3733294}
\acmISBN{979-8-4007-1877-9/2025/06}

\begin{document}

\title{COMAE: COMprehensive Attribute Exploration for Zero-shot Hashing}

\author{Yuqi Li}
\authornote{These authors contributed equally to this work.}
\author{Qingqing Long}
\authornotemark[1]
\email{yuqili010602@gmail.com}
\email{qqlong@cnic.cn}
\affiliation{%
   \institution{CNIC, CAS} 
    \institution{UCAS}
    \city{Beijing}
    \country{China}
}

\author{Yihang Zhou}
\email{yhzhou@cnic.cn}
\affiliation{%
   \institution{CNIC, CAS} 
    \institution{UCAS}
    \city{Beijing}
    \country{China}
}

\author{Ran Zhang}
\author{Zhiyuan Ning}
\email{{zhangran,nzy}@cnic.cn}
\affiliation{%
 \institution{CNIC, CAS} 
    \institution{UCAS}
    \city{Beijing}
    \country{China}
}

\author{Zhihong Zhu}
\email{zhihongzhu@stu.pku.edu.cn}
\affiliation{%
  \institution{Peking University}
    \city{Beijing}
    \country{China}
}

\author{Yuanchun Zhou}
\email{zyc@cnic.cn}
\affiliation{%
\institution{CNIC, CAS} 
    \institution{UCAS}
    \city{Beijing}
    \country{China}
}

\author{Xuezhi Wang}
\authornote{These authors are corresponding authors.}
\author{Meng Xiao}
\authornotemark[2]
\email{{wxz,shaow}@cnic.cn}
\affiliation{%
  \institution{CNIC, CAS} 
    \institution{UCAS}
    \city{Beijing}
    \country{China}
}

\renewcommand{\shortauthors}{Yuqi Li et al.}

\begin{abstract}
Zero-shot hashing (ZSH) has shown excellent success owing to its efficiency and generalization in large-scale retrieval scenarios. However, existing works ignore the locality relationships of representations and attributes, which have effective transferability between seeable classes and unseeable classes. Also, the continuous-value attributes are not fully harnessed. In response, we conduct a \underline{\textbf{COM}}prehensive \underline{\textbf{A}}ttribute \underline{\textbf{E}}xploration for ZSH, named COMAE, which depicts the relationships from seen classes to unseen ones through three meticulously designed explorations, i.e., \textit{point-wise}, \textit{pair-wise}, and \textit{class-wise} consistency constraints. 
By regressing attributes from the proposed attribute prototype network,
COMAE learns the local features that are relevant to the visual attributes. 
Then COMAE utilizes contrastive learning to comprehensively depict the context of attributes, rather than instance-independent optimization. Finally, the class-wise constraint is designed to cohesively learn the hash code, image representation, and visual attributes more effectively. Furthermore, a theoretical analysis is provided to show the effectiveness. Experimental results demonstrate that COMAE outperforms state-of-the-art hashing models, especially in scenarios with a larger number of unseen label classes. Our codes are available at \url{https://github.com/itsnotacie/ICMR2025-COMAE}.
\end{abstract}




\maketitle

\section{Introduction}
Image hashing techniques have garnered substantial interest due to their remarkable efficacy in multimedia retrieval and content-based searches for images and documents \cite{wang2017deep,dong2024pixel,arik2019attention,luo2025large,long2021hgk,li2025wavfusion, li2025prototype, zeng2025enhancing}. Particularly, hashing techniques convert high-dimensional and complicated data into binary codes, which enables efficient searches in the binary Hamming space, thereby contributing to a smaller memory footprint and increased search efficiency \cite{shi2022zero,li2021deep,yang2016zero, guo2017sitnet,zhang2019zero,long2020graph}.

With the popularity of content and social apps, the growth of Internet images leads to sort of new concepts, presenting a challenge for traditional hashing algorithms that struggle to adapt to unseen data in previously unobserved categories \cite{venkatesan2000robust,wang2017deep}. Recent researchers paid attention to the Zero-Shot Hashing (ZSH) issue. Visual attributes, pervasive across prevalent image datasets, delineate discriminative visual properties intrinsic to objects, transcending specific classes and establishing a foundation for addressing the intricacies of previously unseen data  \cite{khan2018evaluating,chen2022transzero, ma2024discrepancy}. Notably, these attributes offer a robust representation for recognizing out-of-distribution samples, making them valuable for capturing similarities in the seen and unseen concepts in Zero-Shot Learning (ZSL) tasks \cite{arik2019attention,xu2017attribute}. 
Nevertheless, the exploration of visual attributes within the realm of ZSH tasks has been relatively meager. AH \cite{xu2017attribute} adopts a straightforward approach by constructing a multi-task architecture to capture the intricate relationships between visual attributes and category classes, thereby facilitating the transfer of knowledge.

\begin{figure*}
\centering
\includegraphics[width=\textwidth]{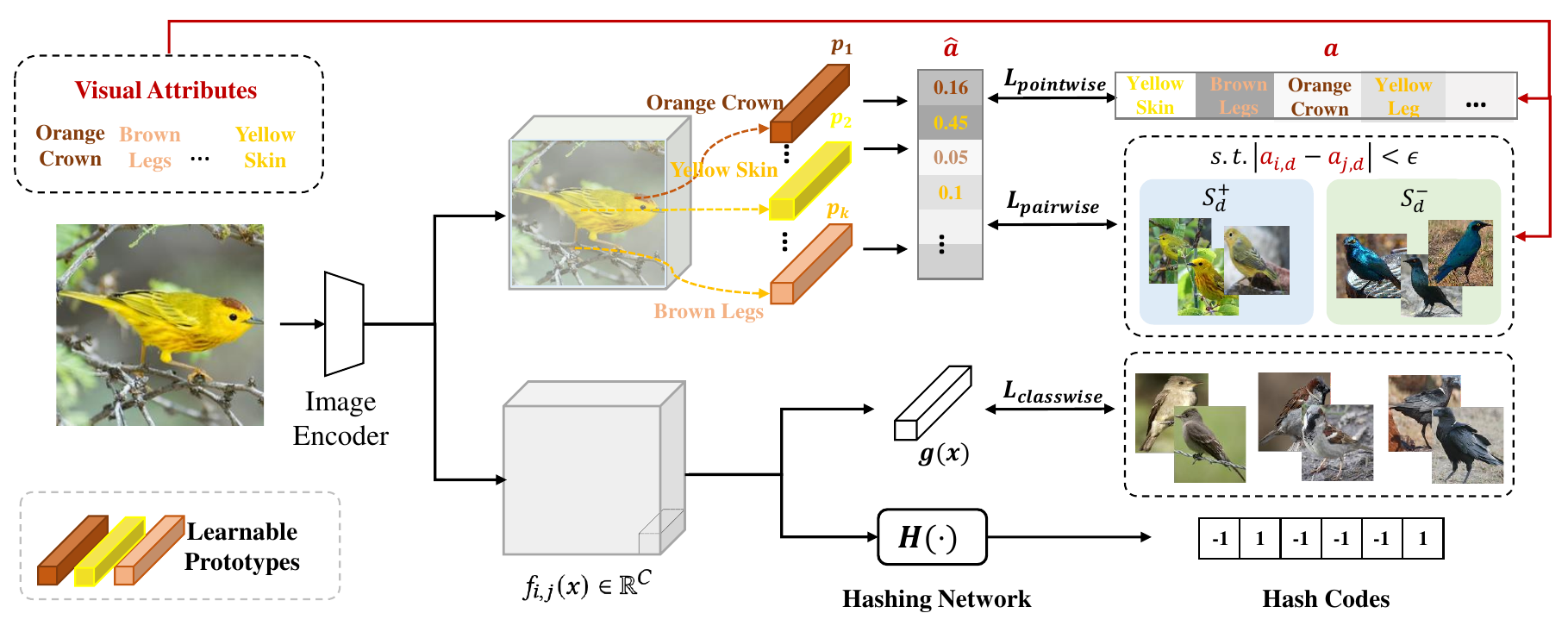}
\vspace{-8mm}
\caption{The architecture of the proposed COMAE. It consists of three modules: a) Point-wise objective aims at improving image locality and attribute representations; b) Pair-wise loss is proposed to learn the representations from individual learning to context-based learning; and c) Class-wise constraint is designed to capture relationships of attributes and class labels.} \label{fig:framework}
% \vspace{-3mm}
\end{figure*}

While considerable success has been achieved in utilizing visual attributes to establish connections among visual attributes, image features, class labels, and binary hash codes, there still exist urgent limitations \cite{li2021deep,xu2017attribute}.
Firstly, image locality representations are proven to be critical to zero-shot learning tasks \cite{li2021adaptive}, as the locality exhibits more effective transferability between seen classes and unseen classes. However, existing works ignore the locality representations in deep hashing tasks, which decreases the upper bound of the transferability in hashing algorithms. 
Secondly, existing works mainly focus on individual image representation learning, optimizing each instance independently. Each dimension of the visual attribute is a continuous value. In contrast to the binary one-hot class encoding, these continuous values encompass richer and more nuanced distance-related information. These informative attributes are aptly poised to capture intricate pairwise relationships, while they are largely ignored by prevailing works.
Thirdly, existing works generally employ a multi-task learning framework to independently utilize attribute labels and class labels. Yet their actual relationships are more complicated than such simple presumptions. To be specific, two images may belong to the same class but pose disparate attributes. For example, as shown in Fig. \ref{fig:framework}, two images with a ``bird'' class may have ``black leg'' and ``brown leg''. Similarly, instances with different classes may also exhibit analogous attributes. For example, two images with ``bird'' and ``wapiti'' classes may have the same ``black leg''.  The existing methods may not lead to the global optimum, thus necessitating performing joint optimization to achieve the overall optimum for both tasks.

To tackle the above issues, we conduct a \textbf{COM}prehensive \textbf{A}ttribute \textbf{E}xploration for zero-shot hashing, named \textbf{COMAE}. 
It combines relationships among visual attributes, image features, class labels, and binary hash codes.  Specifically, we develop a supervised learning objective that localizes visual attributes for point-wise consistency.  By utilizing the attribute prototype network, we aim to capture the fine-grained details and distinctive characteristics of images, enabling more accurate and informative hash codes. 
To comprehensively harness numerical attribute features, we construct positive and negative instances for each image based on the numerical disparities of attributes and utilize contrastive learning to further model their distance relationships. Thus, our model can capture the subtle variations in attribute values across different images and learn representations that encode the common visual characteristics associated with these attributes. 
Finally, class-wise optimization is proposed to cohesively learn the hash code, image representation, and visual attributes more effectively, which fosters improved generalization by considering the relationships and dependencies between different classes during the learning process. 
We jointly learn discriminative global and local features based on the class-level attributes.
Extensive experiments on the ZSH benchmark datasets show that COMAE outperforms the state-of-the-art methods, especially in scenarios with a larger number of unseen label classes. 
Our contributions are summarized as follows:
\begin{itemize}
    \item We conduct a COMprehensive \textbf{A}ttribute \textbf{E}xploration for zero-shot hashing, named COMAE, which depicts the relationships between seen classes and unseen classes through three meticulously designed explorations, i.e., \textit{point-wise}, \textit{pair-wise}, and \textit{class-wise} consistency constraints. 
    \item We design an attribute prototype network, a contrastive learning unit, and a joint-optimal label utilization to implement the above three components, with theoretical analysis to show its effectiveness.  
    \item Experiments on three popular zero-shot hashing datasets show that COMAE significantly outperforms the state-of-the-art methods, especially in scenarios with a larger number of unseen label classes.
\end{itemize}

\section{Related Work}
\subsection{Deep Hashing}
Image hashing is a popularly used method for approximating nearest neighbor search in large-scale retrieval scenarios \cite{venkatesan2000robust, dong2024class,qinapbench,chen2025VSD}. Shallow image hashing methods often leverage handcrafted features or employ traditional machine-learning algorithms to generate compact binary representations. Typical works include LSH \cite{andoni2008near} and ITQ \cite{gong2012iterative}. Shallow methods are computationally efficient and exhibit simplicity in their design, yet they may struggle to capture complex and high-level semantic features inherent in images. 

Over the past decades, deep learning models have shown excellent progress in the field of computer vision \cite{zhu2024multivariate,zhu2024distribution,dong2024pixel,chu2025cross,qin2023destruction}, language processing \cite{wu2024deepseek,guo2025deepseek,cai2023resolving,xiao2025interdisciplinary,lu2024generic}, and multimodal tasks~\cite{lin2024gume,li2021explicit,xiao2021expert,li2024sglp,li2025fedkd}. Deep hashing methods can be achieved through three paths. The first path utilizes limited information, which has a lower upper bound of the capability. ASZH \cite{shi2022zero,qin2022flareon} is a typical work in the first path, which learns hash functions based on the visible category labels, for generating binary codes for unseeable data. The second path \cite{zhang2019zero,guo2017sitnet} generates semantic vectors based on the visual attributes to find a semantic embedding space, then transfers the common knowledge to unseen classes. TSK, SitNet, OPZH, and PIXEL \cite{yang2016zero, guo2017sitnet,zhang2019zero, dong2024pixel} are representative semantic-based works, they conduct a multi-task framework in which the class labels and attribute vectors are co-trained through the hashing network, and the labels are projected into a semantic embedding space. However, PIXEL introduced BERT, which resulted in longer processing times. 
However, there exists an inherent gap between the semantic and visual space, and the alignment of spaces is tough. The third path \cite{xu2017attribute} constructs relationships among attributes, image features, binary codes, and class labels. These models are considered to be more generalized when dealing with out-of-distribution and open-set samples \cite{arik2019attention,gao2019hybrid}.

\subsection{Zero-shot Learning}
Zero-shot learning (ZSL) \cite{farhadi2009describing,xu2020attribute,zheng2022semantic} is designed to extend object recognition across known and unknown classes via a unified embedding space, where both categories are described by their semantic visual identifiers. 
Early ZSL methods \cite{farhadi2009describing,pourpanah2022review,chen2025ambiguity} harvest global visual features from networks, whether pre-trained or capable of end-to-end training. Typically, end-to-end frameworks often surpass pre-trained ones by refining visual attributes, thus mitigating biases between the datasets used for training and those for ZSL applications. Nonetheless,such strategies typically produce less than ideal outcomes \cite{xu2020attribute,li2025PASE}, as they struggle to discern the subtle variances between seen and unseen categories. Recent innovations in ZSL predominantly concentrate on methods based on attention \cite{zheng2022semantic,sylvain2019locality,xu2020attribute}, employing visual attributes to identify distinct feature regions, for instance, specific parts of a bird. This emphasis highlights the critical roles of specificity and the arrangement of image features in ZSL. Proposals for prototype networks \cite{sylvain2019locality} strive to pinpoint various attributes within images to bolster the precision of feature localization. Inspired by these insights into visual attributes and the delicate distinctions between recognized and unrecognized classes, our methodology adopts the prototype network to intricately map the relationships between visual attributes and hash codes \cite{xu2020attribute}.

\section{Model: COMAE}
The architecture of the proposed COMAE is shown in Fig. \ref{fig:framework}. Our proposed COMAE depicts the relationships between seen and unseen classes through three meticulously designed explorations, i.e., point-wise, pair-wise, and class-wise consistency constraints. To improve the image locality and attribute representations, we propose a point-wise objective. To learn the representations from individual learning to context-based learning, we utilize contrastive learning to build a pair-wise loss. For capturing relationships between attributes and class labels, a class-wise constraint is designed.

\subsection{Problem Definition}
Let $X=\{x_i\}_{i=1}^N\in \mathbb{R}^{N\times H\times W\times C}$ be $N$ images, where $H,W,C$ denote the image height, width and channel. The matrix $Y=\{y_i\}_{i=1}^N\in\mathbb{R}^{N\times C}$ denotes the labels, where $C$ denotes the number of classes.
The labels are divided into the seeable class $Y^{s}$ and the unseeable class $Y^{u}$, where $|Y^{s}|\cup |Y^{u}|=|C|, |Y^{s}|\cap |Y^{u}|=\emptyset$. Let $A=\{a_i\}_{i=1}^N\in \mathbb{R}^{N\times K}$ denote the class-level attribute vector  in $K$ dimensions. 
By using training samples which contain labeled images and attributes from seeable classes, i.e., $S=\{x,y,a|x\in X, y\in Y^{s}, a\in A\}$, we aim to learn a hashing model $\mathcal{H}\colon X\rightarrow b_i\in\{-1, 1\}^l$, where $b_i$ denotes learned $l$-bit binary hash codes. For the testing process, attribute vectors of unseen classes are also known. In this paper, zero-shot hashing focuses on predicting the hashing label of images from unseen classes, i.e., $X\rightarrow Y^{u}$.

\subsection{Point-wise Attribute Consistency}
Image locality representations are proven to be practical in transferring knowledge from seen to unseen classes \cite{xu2020attribute,li2021adaptive}. 
For more subtly depicted image and attribute characterizations, we first conduct a point-wise consistency constraint, which takes the attribute regression loss as the supervision. 
Prototype learning has proven to be robust in dealing with open-set recognition and out-of-distribution samples \cite{xu2020attribute,tan2022fedproto} in recent years. In the prototype learning scheme, each category is represented by a prototype. This prototype can be the mean of the sample in the category, the center of mass, or some other representative sample. 
Thus we introduce a prototype module aiming at capturing the relationships of local features and visual attributes. 

Let $f_{i,j}(x)\in\mathbb{R}^C$ denote the image representation at spatial position (or location)  $(i,j)$. It depicts the information on local image regions. Then we learn sorts of attribute prototypes $P=\{p_k\in\mathbb{R}^C\}_{k=1}^K$ to predict attributes from the local visual characterizations. Let $p_k$ denote the learnable prototype embeddings for the $k$-th attribute. As shown in Fig. \ref{fig:framework}, $p_1$ and $p_2$ correspond to the prototypes for ``orange crown'' and ``yellow skin'' respectively. For each attribute, we produce a similarity map $M^k\in\mathbb{R}^{H\times W}$, where each element is computed by a non-linear network $\Theta(\cdot)$ between $p_k$ and each local feature. To associate each visual attribute with its closest local feature, we gain the predicted $k$-th attribute $\hat{a_k}$ by taking the maximum value in the similarity map $M^k$,
\begin{equation}
    \begin{aligned}
    M_{i,j}^k &= \Theta\left (p_k, f_{i,j}(x)\right ),\\
    \hat{a}_k &=\max_{i,j} M_{i,j}^k.
    \end{aligned}
\end{equation}
This prototype captures the common characteristics of the category and plays a key role in the expression of the entire category. The visual attribute vectors supervise the learning of attribute prototypes. We formalize the attribute prediction task as a regression problem and then minimize the Mean Square Error (MSE) between the ground truth attributes and the predicted attribute score $\hat{a}$. By optimizing the regression objective, we enforce the local features to encode pivotal semantic attributes, improving the locality of representations, and further improving the quality of hashing codes,
\begin{equation}
     \mathcal{L}_{\text{pointwise}} =\sum_i^{N}\frac{1}{N}||a_i-\hat{a_i}||^2.
     \label{eq:pointwise}
\end{equation}

\subsection{Pair-wise Attribute Consistency}
To make full use of the limited yet crucial pieces of information, we conduct pair-wise attribute exploration to depict the context relationships of continuous-valued attributes. Recent studies proved that contrastive learning can enhance the discriminative capability of feature representations and enable the establishment of meaningful relationships between seen and unseen classes \cite{han2021contrastive,qiu2021unsupervised,chen2018contrastive}. The fundamental concept of pairwise loss revolves around learning feature representations by comparing the similarity between different pairs of samples.  The pairwise loss function enhances the model's ability to compare sample pairs across these groups, directing it to recognize that samples within the same category should exhibit higher similarity compared to those from different categories. This approach ensures a more discerning and category-aware similarity assessment within the model's learning process.
Thus we utilize contrastive learning in the pair-wise constraint.
We begin with constructing the positive sample set $S^{+}_d(i)$ of the $i$-th image and the negative sample set $S^{-}_d(i)$ in the $d$-th dimension of all attributes. Specifically, given the attribute of $i$-th image, which denoted as $a_i$, we build the positive set of $i$-th image as follows,
\begin{equation}
    S^{+}_d(i)=\{j\mid \lvert a_{i,d} -  a_{j,d}\rvert < \epsilon,  j\neq i, j=1,...,N \},
    \label{eq:attribute_positive_sample}
\end{equation}
where $\epsilon$ is a hyperparameter and is set through multiple experiments. We randomly sample $k$ negative samples for the $i$-th image, as considering all negative samples will bring high complexity. We get the sample set $S_d(i)=S^{+}_d(i)\cup S^{-}_d(i)$.
Afterwards, we formulate the attribute-level contrastive embedding objective as follows,
\begin{equation}
   \mathcal{L}_{\text{pairwise}} = -\sum_d^D\sum_i^N\frac{1}{D N}
   \log \frac{\sum_{j\in S^{+}_d(i)}\text{exp}(\hat{a_i}\hat{a_j}/\tau)}{\sum_{t\in S_d(i)}\text{exp}(\hat{a_i}\hat{a_t}/\tau)} ,
   \label{eq:pairwise}
\end{equation}
where $\tau > 0$ is the temperature parameter. The attribute-level contrastive learning relies on attribute-wise supervision to enhance the discriminative ability in the new attribute space.

\begin{algorithm}[t]
\caption{Algorithm of COMAE}
\label{alg:COMAE}
\begin{algorithmic}[1]
\Require Training Images $X=\{x_i\}_{i=1}^N$, class labels $Y=\{y_i\}_{i=1}^N$, and image attribute vectors $A=\{a_i\}_{i=1}^N$.
\Ensure The parameters $\Theta$ 
\For{$i=0,..., N$}
\State Generate image initial embeddings $f(x_i)$ based on the image encoder, i.e., ResNet101;
\EndFor
\State Generate the predicted prototypes $\{M^k\}_{k=1}^K$ for each attribute.
\For{$i=0,..., N$}
    \State Construct the positive sample set $S^{+}_d(i)$ of the $i$-th image and negative sample set $S^{-}_d(i)$, as Eqn. (\ref{eq:attribute_positive_sample}).
\EndFor
\While {not converge} :
    \State Compute the pointwise loss, as Eqn. (\ref{eq:pointwise}).
    \State Compute the pairwise loss, as Eqn. (\ref{eq:pairwise}).
    \State Compute the classwise loss, as Eqn. (\ref{eq:classwise}).
    \For{$i=0,..., N$}
        \State Generate binary hashing codes $B=\{b_i\}_{i=1}^N$ based on image  representations $f(x_i)$ through Eqn. (\ref{eq:hash_layer}).
    \EndFor
    \State Compute the joint loss through Eqn. (\ref{eq:total_loss})
    \State  Compute the backpropagation error.
    \State Update the parameters $\Theta$ and image hidden representations $f(x_i)$.
\EndWhile \\

\Return Image hashing codes $B=\{b_i\}_{i=1}^N, b_i\in\{-1, 1\}^l$.
\end{algorithmic}
\end{algorithm}

\subsection{Class-wise Attribute Exploration}
Class-wise attribute exploration aims to leverage image attributes to enhance the performance of classification. Compared with pointwise and pairwise explorations, it focuses on higher-level relationships of images and attributes. For high-level image representations, we adopt a global average pooling over the input image to learn a global discriminative feature $g(x)\in \mathbb{R}^C$, $g(x) = \frac{1}{H W}\sum_{i=1}^H\sum_{j=1}^W f_{i,j}(x)$. 
Subsequently, a linear layer employing parameter $W\in \mathbb{R}^{C\times K}$ maps the visual feature $g(x)$ into the designated class embedding space (e.g., attribute space). Calculating the dot product between this transformed visual feature and each class embedding generates class logits. This process is complemented by applying a cross-entropy loss function, which is designed to maximize the compatibility score between the image and its corresponding attribute vector, fostering a more accurate attribute-based classification,
\begin{equation}
    \mathcal{L}_{\text{ce}} = -\frac{1}{|Y^s|}\sum_{y_i\in Y^s}\log  \frac{\text{exp}\left (g(x_i)^T W a_i\right )}{\text{exp}\left (g(x_i)^T W  \hat{a_i}\right )} ,
    \label{eq:classwise}
\end{equation}
The attribute embedding layer and image representations are jointly optimized, with class labels directing the overall optimization effort. Note that the attribute prototypes, which facilitate the transfer of knowledge from seen to unseen classes, are shared across different classes. This sharing mechanism enhances image representation for zero-shot learning, improving the model's ability to generalize to new, unseen classes.

\subsection{Hashing Learning}
After conducting the above three consistency constraints, the image representations can contain more comprehensive attribute and class relationships. In this section, we derive a hash network based on the improved representations.
Denote our hashing network as $H({\cdot})$, given an input image representation $f(x)$ from the pre-trained image encoder $f({\cdot})$. We take the Fully Connected Hash (FCH) as the fully connected layer in $f({\cdot})$ for generating binary code $b_i\in\mathbb{R}^d$, by binarization process with the use of $\text{sign}$ function. The process is formulated as follows,
\begin{equation}
b_i=\text{sign}\left (\sum_{c=1}^C\sum_{h=1}^H\sum_{w=1}^Wf(x_i)W_{h,w,c}\right),
\label{eq:hash_layer}
\end{equation}
where $W_{h,w,c}$ is the learnable transform matrix. It changes the last fully connected layer to a $d$-dimension representation.
To compress intra-class hashing code with strong constraints, we use the hypersphere loss \cite{liu2017sphereface,wang2018additive}  to optimize the FCH layer with large angular boundaries between classes as $\mathcal{L}_{\text{hash}}$. Compared with other hashing methods, it can achieve a larger separation and a smaller variation of inter-classes.

For a comprehensive exploration between the attribute learning and hashing tasks, the \textbf{overall objective} is formalized as follows,
\begin{equation}
\mathcal{L} = \sum\lambda_i\mathcal{L}_i ,
\label{eq:total_loss}
\end{equation}
where $\mathcal{L}_i$ denotes the pointwise-, pairwise-, classwise-, and hashing losses respectively, and $ \lambda_i$ denotes hyper-parameters. The parameters of the hashing network could not be updated by the standard back-propagation as the derivation of $\text{sign}(\cdot)$ is indifferentiable at zero. Thus $\text{tanh}(\cdot)$ is adopted to approximate the results of $\text{sign}(\cdot)$. In this manner, we can optimize the original non-differentiable loss by the mini-batch Stochastic Gradient Descent (SGD). 
The pseudocode of COMAE is shown in the Algorithm. \ref{alg:COMAE}.

\subsection{Theoretical Analysis}
Here we provide a theoretical analysis for our proposed COMAE. 
\begin{theorem}\cite{zhu2022lower} Let $\mathcal{D}_{\mathit{inter}}$ and $\mathcal{D}_{\mathit{intra}}$ denote the \textit{inter-class distinctiveness} and \textit{intra-class compactness}, respectively.  The lower bound of deep supervised hashing performance is proportional to
\begin{equation}
    \frac{\min{\mathcal{D}_{\mathit{inter}}}}{\max{\mathcal{D}_{\mathit{intra}}}},
    \quad\mathit{s.t.}\; \bm{\mathcal{X}}  \xrightarrow{\mathcal{H}(\cdot)} \bm{\mathcal{B}} . 
\end{equation}
\end{theorem}
Then we analyze the effectiveness of our proposed COMAE. Let $a_c$ denote the attribute vector associated with class $c$, and  $\mathcal{D}^{\mathit{Attr}}_{\mathit{inter}}=||a^{c_1}_i-a^{c_2}_i||, c_1\neq c_2$ denote the distance between attribute vectors of any two distinct classes. Let $\mathcal{D}^{\mathit{Attr}}_{\mathit{intra}}=||b_i-\tau(c_i)||$ denote the maximum distance from any sample within a class to its class center, where $\tau(c)$ is the center of class $c$.
Following the above theorem, we have the following conclusion in our attribute-based zero-shot hashing settings,
\begin{theorem}  The performance of attribute zero-shot hashing depends on $ \frac{\min{\mathcal{D}^{\mathit{Attr}}_{\mathit{inter}}}}{\max{\mathcal{D}^{\mathit{Attr}}_{\mathit{intra}}}}$.
By optimizing for higher $\mathcal{D}^{\mathit{Attr}}_{\mathit{inter}}$ and lower $\mathcal{D}^{\mathit{Attr}}_{\mathit{intra}}$, the lower bound expressivity of attribute zero-shot hashing can be improved.
\end{theorem}
The pairwise loss $\mathcal{L}_{\text{pairwise}}$ (Eqn. \ref{eq:pairwise}) in COMAE can be viewed as $\mathcal{D}^{\mathit{Attr}}_{\mathit{inter}}$ and our class-wise loss $\mathcal{L}_{\text{ce}}$ (Eqn. \ref{eq:classwise}) can be viewed as  $\mathcal{D}^{\mathit{Attr}}_{\mathit{intra}}$. They together improve the performance of the lower bound and indicates that COMAE has better theoretical performance.

\begin{table*}
  \centering
  \caption{The mAP@5000 results on three popular ZSH datasets in different code lengths.}
  \resizebox{\linewidth}{!}{
    \begin{tabular}{cccccccccccccc}
    \toprule
    & \multicolumn{6}{c}{AWA2} & \multicolumn{2}{c}{CUB} & \multicolumn{5}{c}{SUN} \\
    \cmidrule(lr){3-6} \cmidrule(lr){7-10}  \cmidrule(lr){11-14} 
    Methods & Reference  & 24 bits  & 48 bits  & 64 bits  & 128 bits & 24 bits & 48 bits  & 64 bits  & 128 bits & 24 bits & 48 bits  & 64 bits  & 128 bits \\
    \midrule
    LSH & CACM08 & 0.0106 & 0.0151 & 0.0204 & 0.0306 & 0.0055 & 0.0069 & 0.0076 & 0.0095 & 0.0598 & 0.0738 & 0.0764 & 0.0880\\
    SH & NeurIPS08 & 0.1833 & 0.2729 & 0.2955 & 0.3441 & 0.0568 & 0.081 & 0.0886 & 0.1191 & 0.0738 & 0.0811 & 0.0837 & 0.0880 \\
    ITQ & TPAMI12 & 0.1999 & 0.2821 & 0.2964 & 0.3764 & 0.0533 & 0.0765 & 0.0892 & 0.1182 & 0.0725 & 0.0899 & 0.0918 & 0.1007\\
    IMH & TIP15 & 0.1282 & 0.1536 & 0.1613 & 0.1681 & 0.0330 & 0.0361 & 0.0364 & 0.0386 & 0.0572 & 0.0668 & 0.0638 & 0.0651 \\
    PCA & CG93 & 0.2165 & 0.2530 & 0.2701 & 0.2719 & 0.0547 & 0.0598 & 0.0632 & 0.0695 & 0.0808 & 0.0868 & 0.0891 & 0.0980\\
    \midrule
    HashNet  & ICCV17   &0.2086   &0.2386    &0.2516    &0.2749    &0.0528   &0.0566    &0.0595    &0.0633    &0.0746   &0.0825    &0.0865    &0.0944\\
    GreedyHash & NeurIPS18 &0.3420   &0.4169    &0.4240     &0.4639     &0.1132  &0.1707   &0.1841     &0.2326    &0.0873   &0.1259    &0.1398    &0.1550\\
    JMLH    & ICCV19    &0.3607   &0.4364    &0.4408    &0.4711     &0.1078  &0.1555   &0.1987     &0.2310    &0.0860   &0.1168    &0.1345    &0.1500\\
    ADSH    & Access19    &0.3360   &0.4787    &0.5105    &0.5454     &0.0858  &0.1607    & 0.1827   & 0.2424   &0.0886    &0.1230    &0.1341    &0.1731\\ 
    CSQ   & CVPR20   &0.3194    &0.3988    &0.3773    &0.4072      &0.0996    &0.1588    &0.1712     &0.2201   &0.0915  &0.1211  &0.1375  &0.1669\\
    DPN    & IJCAI20   &0.1783  &0.2086    &0.2378    &0.2565      &0.0445    &0.0728    &0.0772      &0.1003   &0.0737  &0.0825  &0.0894  &0.1126\\ 
    BiHalf  & AAAI21 & 0.3440    &0.4036   &0.4223    &0.4577     &0.0794    &0.1280    &0.1573      &0.2142   &0.0077  &0.0490  &0.0512  &0.1307\\
    OrthoCos & NeurIPS21 &0.1709    &0.2302    &0.2312    &0.2566     &0.0451   &0.0660    &0.0736    &0.0984     &0.0728  &0.0809   &0.0868    &0.1126\\
    CIBHash & IJCAI21  &0.2113   &0.2304   &0.2481    &0.2618      &0.0351    &0.0404     &0.0411    &0.0453     &0.0690  &0.0808   &0.0809   &0.0883\\ 
    TBH  & CVPR20    &0.0941   &0.1201  &0.1073    &0.1730      &0.0157  &0.0176     &0.0226       &0.0252    &0.0169  &0.0230  &0.0248  &0.0287\\

     \midrule
     TSK & MM16 & 0.2262 & 0.3109 & 0.3873 & 0.4151 & 0.0739 & 0.1200 & 0.1394 & 0.1112 & \\
     ASZH & TKDE22 & 0.2619 & 0.3787 & 0.4032 & 0.4158 & 0.0764 & 0.1192 & 0.1294 & 0.1727 & - & - & -& -\\ 
    SitNet & IJCAI17 & 0.2344  &  0.2406  &  0.2549 &   0.2650  &  0.0880  &  0.1127  &  0.1141 &   0.1167 & - & - & -& - \\ 
    OPZH & PRL19 & 0.1056 &  0.1390 &  0.1618 &  0.1961 &  0.0632 &  0.0879  & 0.0962 &  0.1143 & - & - & -& -\\ 
    AH & ICME17  & 0.2275  & 0.1989  & 0.3154  & 0.3557  & 0.0480  & 0.0897  & 0.1089  & 0.1445 & - & - & -& -\\ 
    PIXEL & CIKM24 &\textbf{0.3981} &0.4792  & 0.5133  &0.5465  &0.1136  &0.1777 &0.1994  &0.2519   &0.0923  &0.1264  &0.1415  &0.1741 \\
    \midrule
    \textbf{COMAE} & Ours & 0.3843 & \textbf{0.5073} & \textbf{0.5149} &   \textbf{0.5801} &   \textbf{0.1142} &   \textbf{0.1803} & \textbf{0.2012}  &\textbf{0.2532} 
    &\textbf{0.0925}  &\textbf{0.1278} &\textbf{0.1465}  &\textbf{0.1764}\\
    \bottomrule
    \end{tabular}%
    } %resizebox
\label{tab:res1}
\end{table*}

\section{Experiments}
In this section, we will outline the experimental setups and provide a detailed discussion of our results. Our experiments are designed to answer the following questions (\textbf{RQ}): 

\begin{itemize}
    \item \textbf{RQ1:}
    Can COMAE gain superior performance in comparison to other baselines?
    \item \textbf{RQ2:} How does COMAE perform in response to the change in the ratio of unseen classes in the training process?
    \item \textbf{RQ3:} How does COMAE perform in case studies, where its effectiveness is evaluated in a more intuitive way?
    \item \textbf{RQ4:} Is COMAE a running efficient hashing model?
    \item \textbf{RQ5:} How does COMAE perform in model analysis, including the evaluation of its component change?
\end{itemize}

\subsection{Experimental Settings}
 
\subsubsection{Datasets}
In this paper, we analyze three distinguished zero-shot hashing datasets referenced in \cite{shi2022zero}, i.e., AWA2 \cite{xian2018zero},  CUB \cite{wah2011caltech}, SUN \cite{patterson2012sun}. AWA2 encompasses 37,322 images across 50 animal categories, with a configuration of 40 seen and 10 unseen classes, each described by 85 attributes [39]. The CUB dataset includes 11,788 images from 200 bird species, categorized into 150 seen and 50 unseen classes, each characterized by 312 attributes. SUN dataset, on the other hand, contains 14,340 images from 717 scene categories, with a division of 645 seen and 72 unseen classes, each depicted by 102 attributes. 
\subsubsection{Baselines}
We select the following representative image hashing baselines as our competitors:

\noindent{\textbf{(1) Image Shallow Hashing} methods often leverage handcrafted features or employ traditional machine-learning algorithms to generate compact binary representations. Typical works include LSH \cite{andoni2008near}, SH \cite{weiss2008spectral}, ITQ \cite{gong2012iterative}, IMH \cite{shen2015hashing}, and PCA \cite{mackiewicz1993principal}. Shallow methods are computationally efficient and exhibit simplicity in their design, yet they may struggle to capture complex and high-level semantic features inherent in images.}

\noindent{\textbf{(2) Image Deep Hashing} methods, leverage deep neural networks to automatically learn hierarchical and abstract representations of images. Representative works can be categorized into two paths, supervised hashing works, i.e,  HashNet \cite{cao2017hashnet}, GreedyHash \cite{su2018greedy}, JMLH \cite{shen2019embarrassingly}, ADSH \cite{zhou2019angular}, CSQ \cite{yuan2020central}, DPN \cite{fan2020deep}, OrthoCos \cite{hoe2021one}, and unsupervised hashing works, i.e, BiHalf \cite{li2021deep}, TBH \cite{shen2020auto}, CIBHash \cite{qiu2021unsupervised}. By employing end-to-end learning, deep hashing methods aim to optimize hash codes directly from raw pixel values, allowing for the discovery of intricate and discriminative image representations. }

\noindent{\textbf{(3) Zero-shot Image Hashing} methods extend the above algorithms to handle scenarios where the model encounters classes during testing that were not seen during training. These methods often incorporate semantic information, such as image attribute descriptions, or meta-learning strategies to enhance the model's adaptability to unseen classes. Representative works include TSK \cite{yang2016zero}, ASZH, SitNet, OPZH, and AH, PIXEL\cite{dong2024pixel}. ASZH \cite{shi2022zero} uses category labels to learn the hash functions from the seen training data, to generate binary codes for unseen data. SitNet \cite{guo2017sitnet}, and OPZH \cite{zhang2019zero} adopt semantic vectors to find a semantic embedding space for knowledge transferability. AH \cite{xu2017attribute} constructs relationships and image features for transferability. PIXEL\cite{dong2024pixel} improves accuracy by incorporating BERT as the text encoder.}

\subsubsection{Evaluation Metric}
To evaluate different aspects of our study, we employed a variety of metrics. These include the mean Average Precision (mAP) and the Area Under the Precision-Recall Curve (AUC) for assessing the ranking quality of search outcomes. Specifically, we present findings for the top 5,000 samples and for the entire dataset, denoted as mAP@5000 and mAP@all, respectively. Additionally, we utilized four conventional metrics: mean Average Precision (mAP), Precision-Recall Curve (PR Curve), Precision@N Curve (P@N Curve), and Recall@N Curve (R@N Curve), with a particular focus on the ranking performance for the initial 5,000 retrieved samples.

\subsubsection{Implementation Details}
All comparative baselines are conducted with their default hyper-parameter configurations. For a fair comparison, their image feature extractors are substituted with the ResNet101 \cite{khan2018evaluating}. The training epoch is set to 10 and the batch size is set to 64, more than 10 will be overfitting, resulting in a decrease in accuracy. We take Adam as the optimizer weight decay 0.0005, and the learning rate is set to 0.0001. The $\epsilon$ in Eqn. \ref{eq:attribute_positive_sample} is set to 0.9, the $\lambda_1,...,\lambda_4$ in Eqn. \ref{eq:attribute_positive_sample} are set to 10, 1, 10, and 1, respectively.
To ensure the statistical significance of our experimental data, the average outcomes of each technique were calculated based on ten iterations.

\begin{figure}
\centering
\subfigure[PR Curve (AWA2)]{\includegraphics[width=0.31\linewidth]{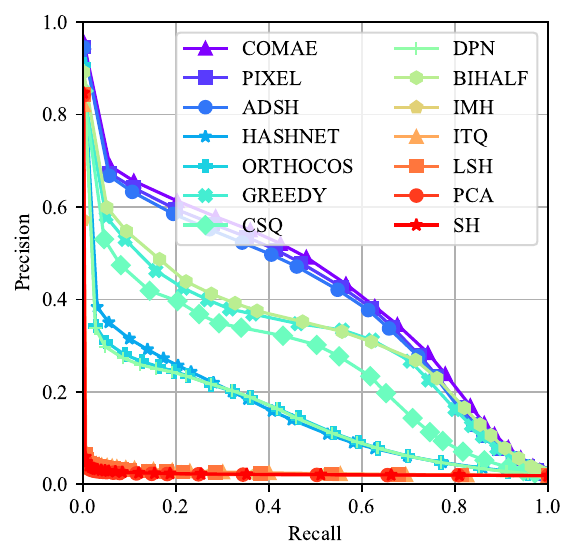} }
\subfigure[PR Curve (CUB)]{\includegraphics[width=0.31\linewidth]{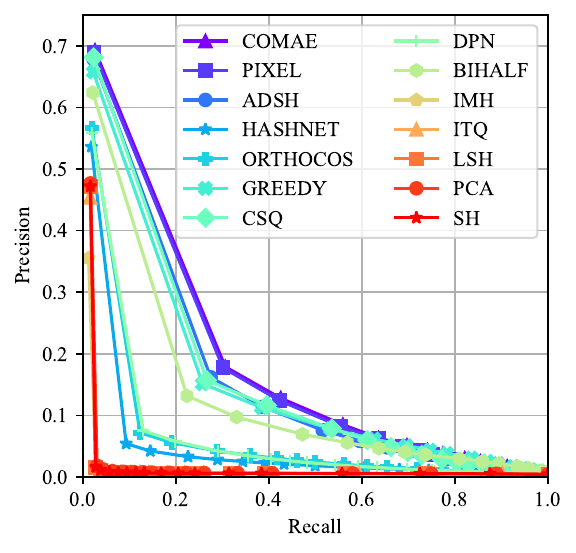} }
\subfigure[PR Curve (SUN)]{\includegraphics[width=0.31\linewidth]{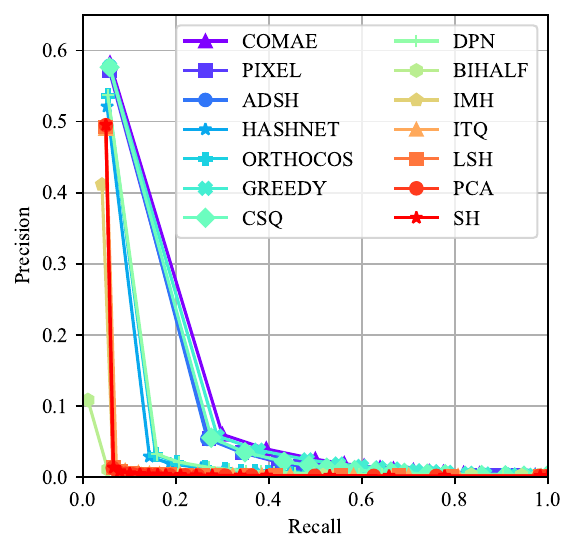} }
\subfigure[P@N Curve (AWA2)]{\includegraphics[width=0.31\linewidth]{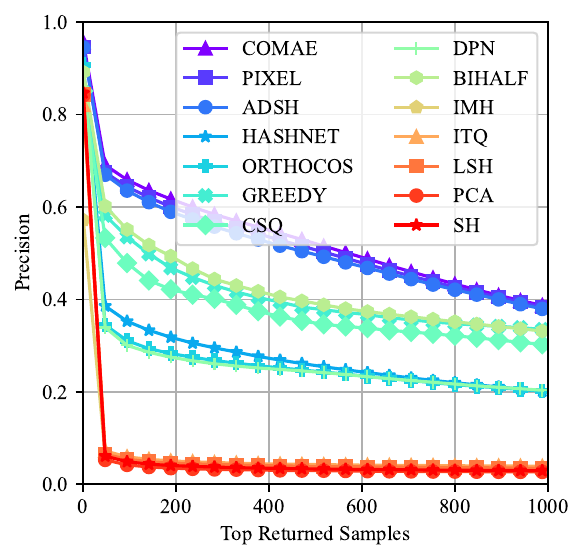}}
\subfigure[P@N Curve (CUB)]{\includegraphics[width=0.31\linewidth]{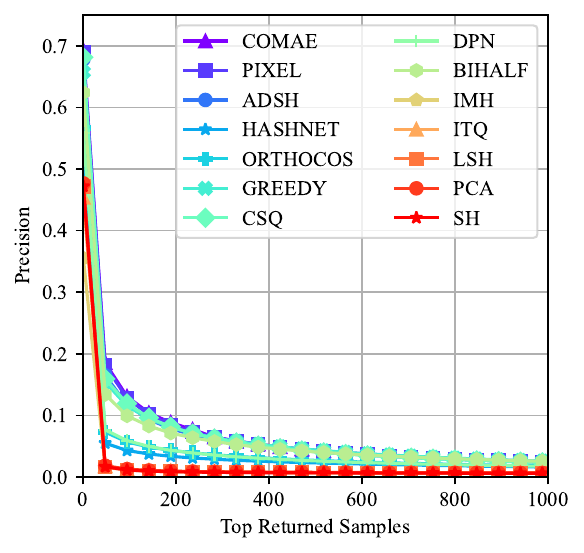}}
\subfigure[P@N Curve (SUN)]{\includegraphics[width=0.31\linewidth]{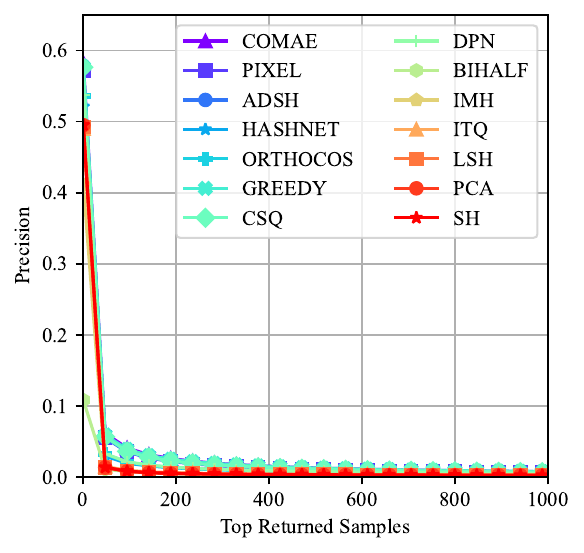}}
\subfigure[R@N Curve (AWA2)]{\includegraphics[width=0.31\linewidth]{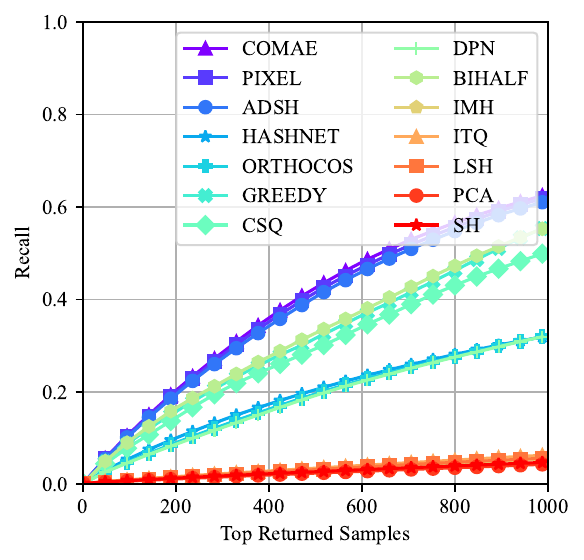}}
\subfigure[R@N Curve (CUB)]{\includegraphics[width=0.31\linewidth]{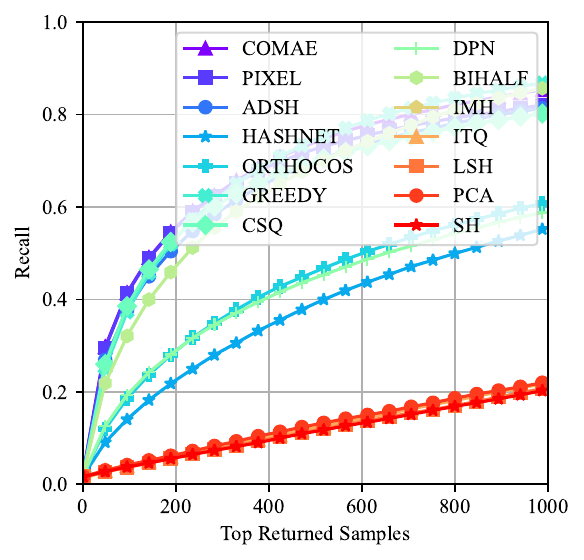}}
\subfigure[R@N Curve (SUN)]{\includegraphics[width=0.31\linewidth]{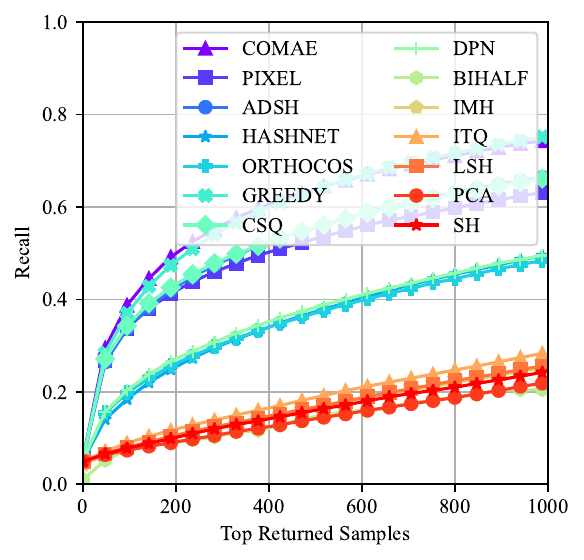}}
\caption{The comparison of PR Curve, P@N Curve and R@N Curve in 64 bit length codes.} 
% \vspace{-3mm}
\label{fig:curve_pr}
\end{figure}

\subsection{RQ1: Overall Performance}
\begin{figure}[!t]
    \centering
    \subfigure[mAP@all in AWA2]{\includegraphics[width=0.3\linewidth]{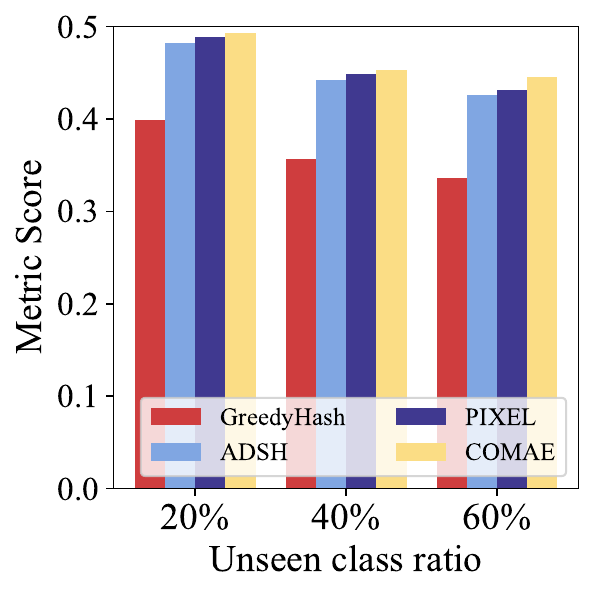}
    }
    \subfigure[mAP@all in CUB]{\includegraphics[width=0.3\linewidth]{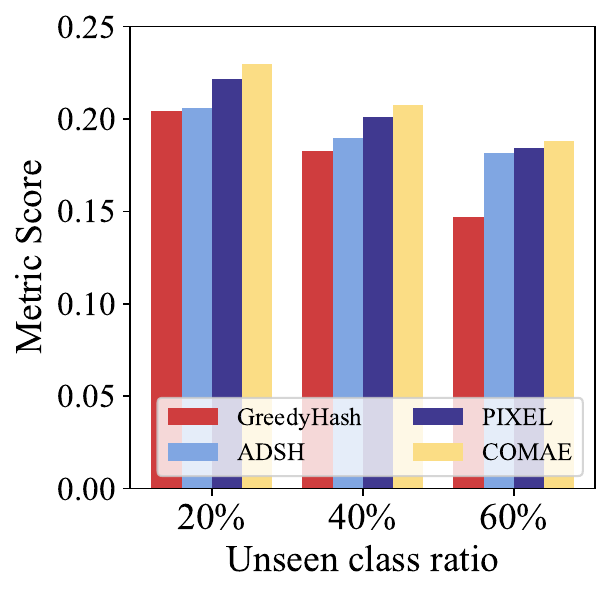}
    }
    \subfigure[mAP@all in SUN]{\includegraphics[width=0.3\linewidth]{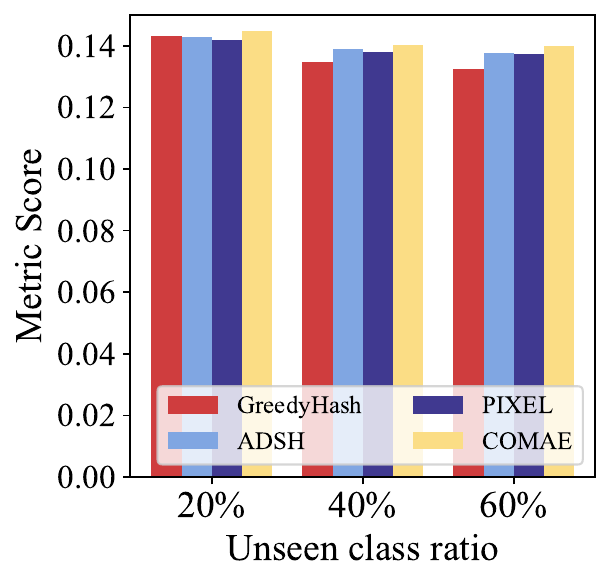}
    } \\
    \subfigure[Decline Percentage (mAP) in AWA2]{\includegraphics[width=0.3\linewidth]{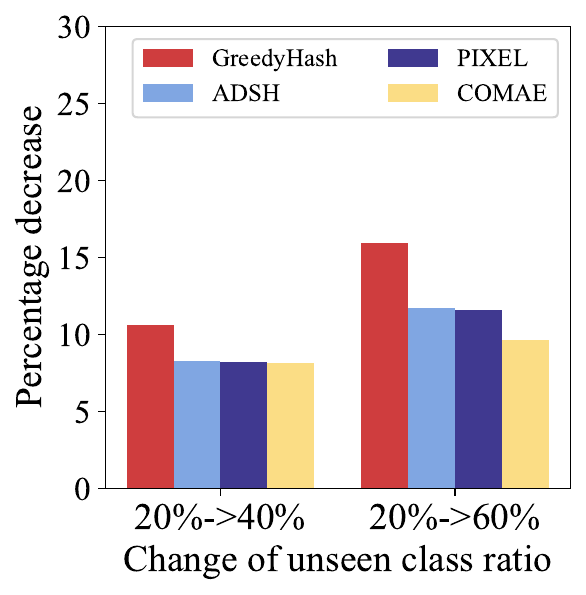}
    }
    \subfigure[Decline Percentage (mAP) in CUB]{\includegraphics[width=0.3\linewidth]{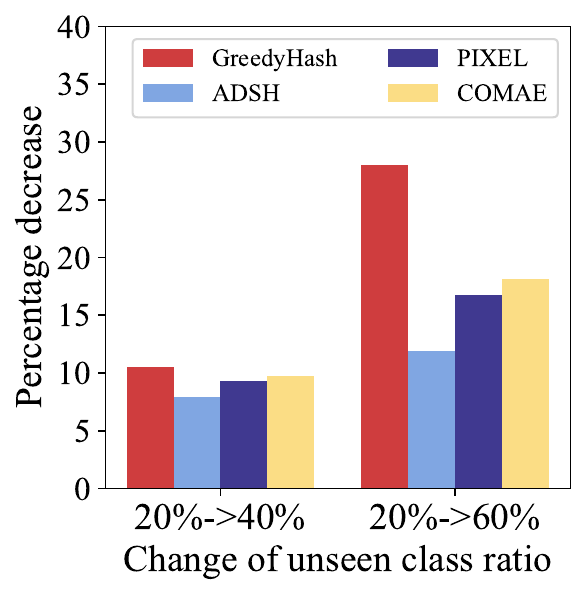}
    }
    \subfigure[Decline Percentage (mAP) in SUN]{\includegraphics[width=0.3\linewidth]{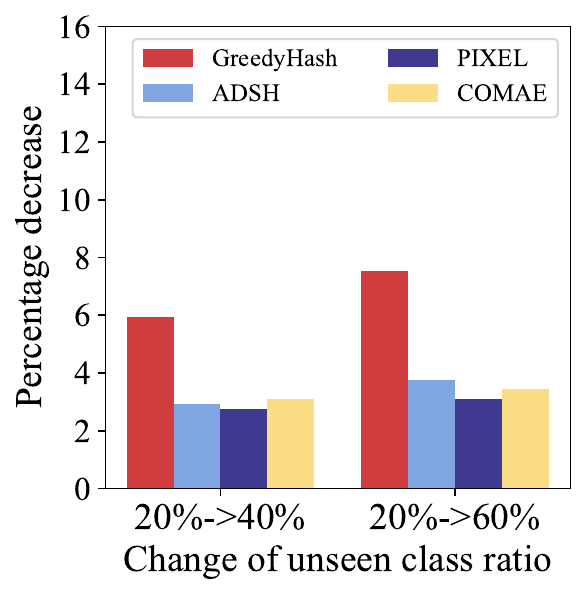}
    }\\
    \subfigure[AUC in AWA2]{\includegraphics[width=0.3\linewidth]{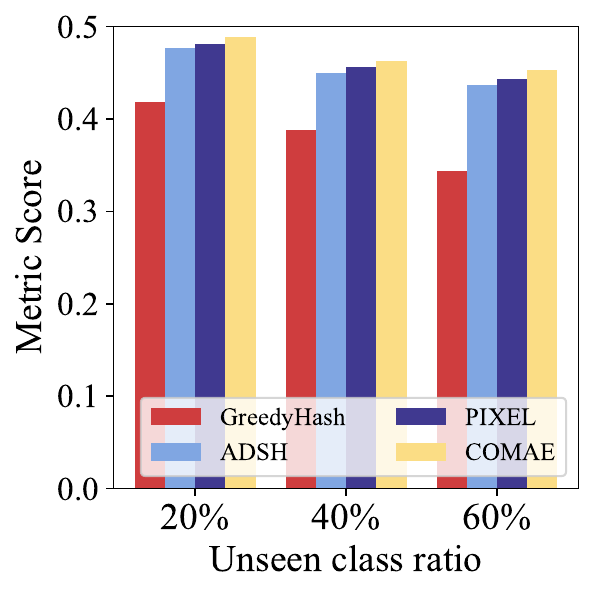}
    }
    \subfigure[AUC in CUB]{\includegraphics[width=0.3\linewidth]{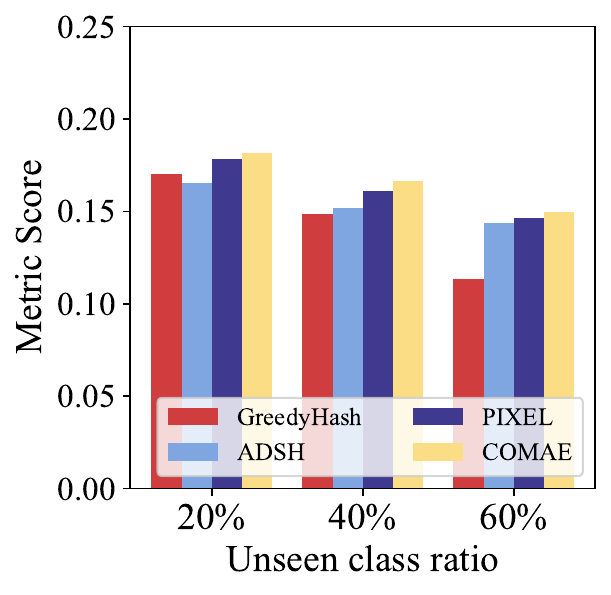}
    }
    \subfigure[AUC in SUN]{\includegraphics[width=0.3\linewidth]{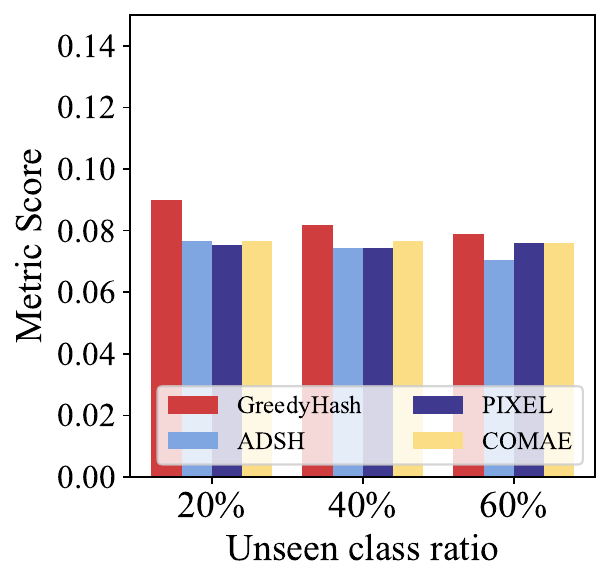}
    }\\
    \subfigure[Decline Percentage (AUC)  in AWA2]{\includegraphics[width=0.3\linewidth]{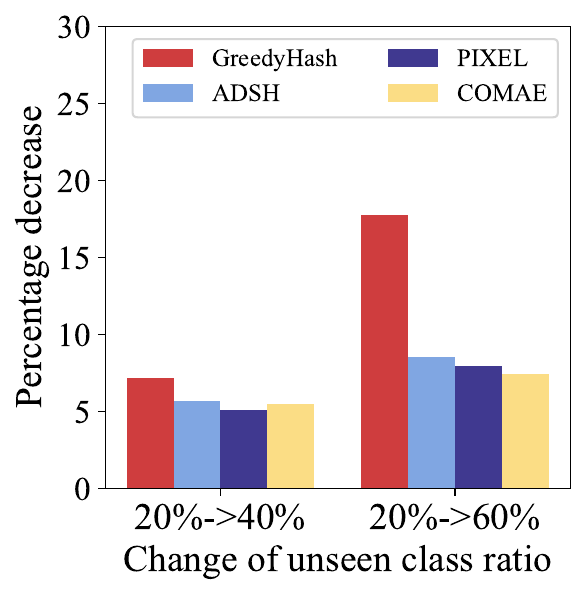}
    }
    \subfigure[Decline Percentage (AUC)  in CUB]{\includegraphics[width=0.3\linewidth]{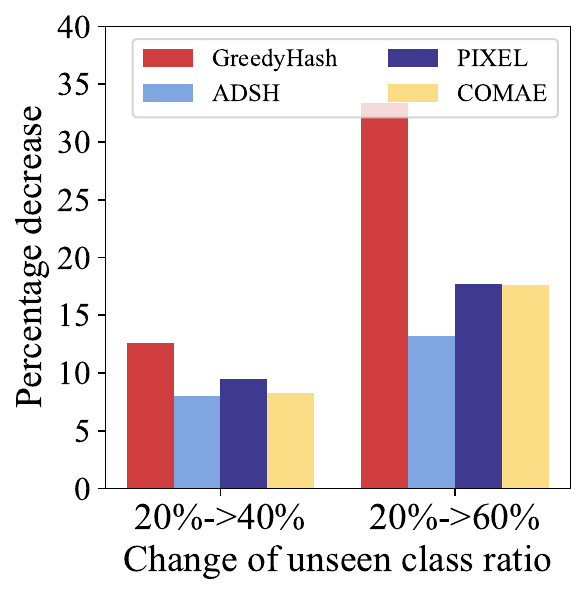}
    }
    \subfigure[Decline Percentage (AUC)  in SUN]{\includegraphics[width=0.3\linewidth]{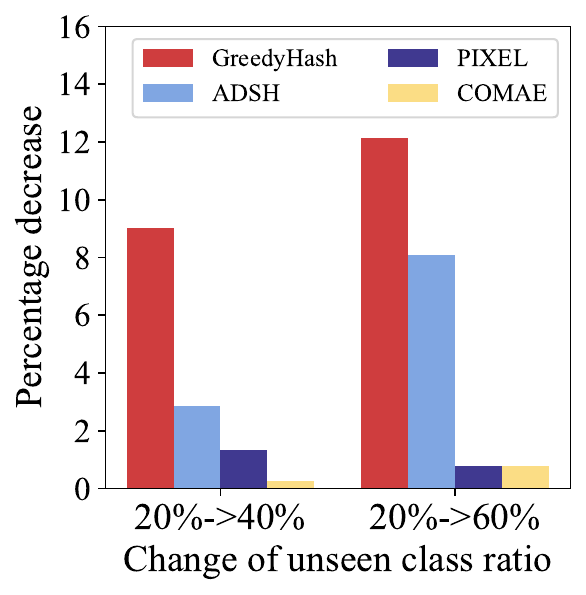}
    }
    \caption{mAP and AUC scores with the change of unseen classes ratio in the training process.}
    % \vspace{-3mm}
    \label{fig:zero_ratio}
\end{figure}

Table. \ref{tab:res1} displays the results of the COMAE and all competitive baselines on datasets AWA2, CUB, and SUN with hash code lengths of 24, 48, 64, and 128. Note that the source codes of ASZH, SitNet, AH, and OPZH are unavailable, thus we use their results in the paper \cite{shi2022zero}. The experimental settings of Table \ref{tab:res1} are the same as \cite{shi2022zero}.  
In Fig. \ref{fig:curve_pr}, we illustrate three distinct curves, i.e., PR Curve, R@N Curve, and the P@N Curve, to evaluate the performance of hash codes with a length of 64 bits on all datasets. Typically, the efficacy of a method is gauged by the PR curve approaching the upper right quadrant, with higher positions on the P@N and R@N curves indicative of superior performance. 

We find the following conclusions: \textbf{(1)} As shown in Table. \ref{tab:res1}, deep hashing models perform better than shallow hashing ones in general, which shows the superiority of neural networks. In addition, the zero-shot hashing models exhibit superior performance in comparison to both deep and shallow hashing models. \textbf{(2)} COMAE surpasses all existing zero-shot hashing benchmarks, in terms of all bit lengths. At the same time, the performance of COMAE remains relatively stable when the number of bits increases. Compared with PIXEL, the latest attribute-based zero-shot hashing method, we gain better performance, which serves to validate the correctness and effectiveness of the motivation we put forth. \textbf{(3)} Based on the results shown in Fig. \ref{fig:curve_pr} across all datasets, COMAE attains a heightened accuracy at a comparatively lower recall rate, thereby demonstrating its superiority.

\begin{figure}[!t]
    \centering
    \subfigure[without $\mathcal{L}_\text{classwise}$]{\includegraphics[width=0.4\linewidth]{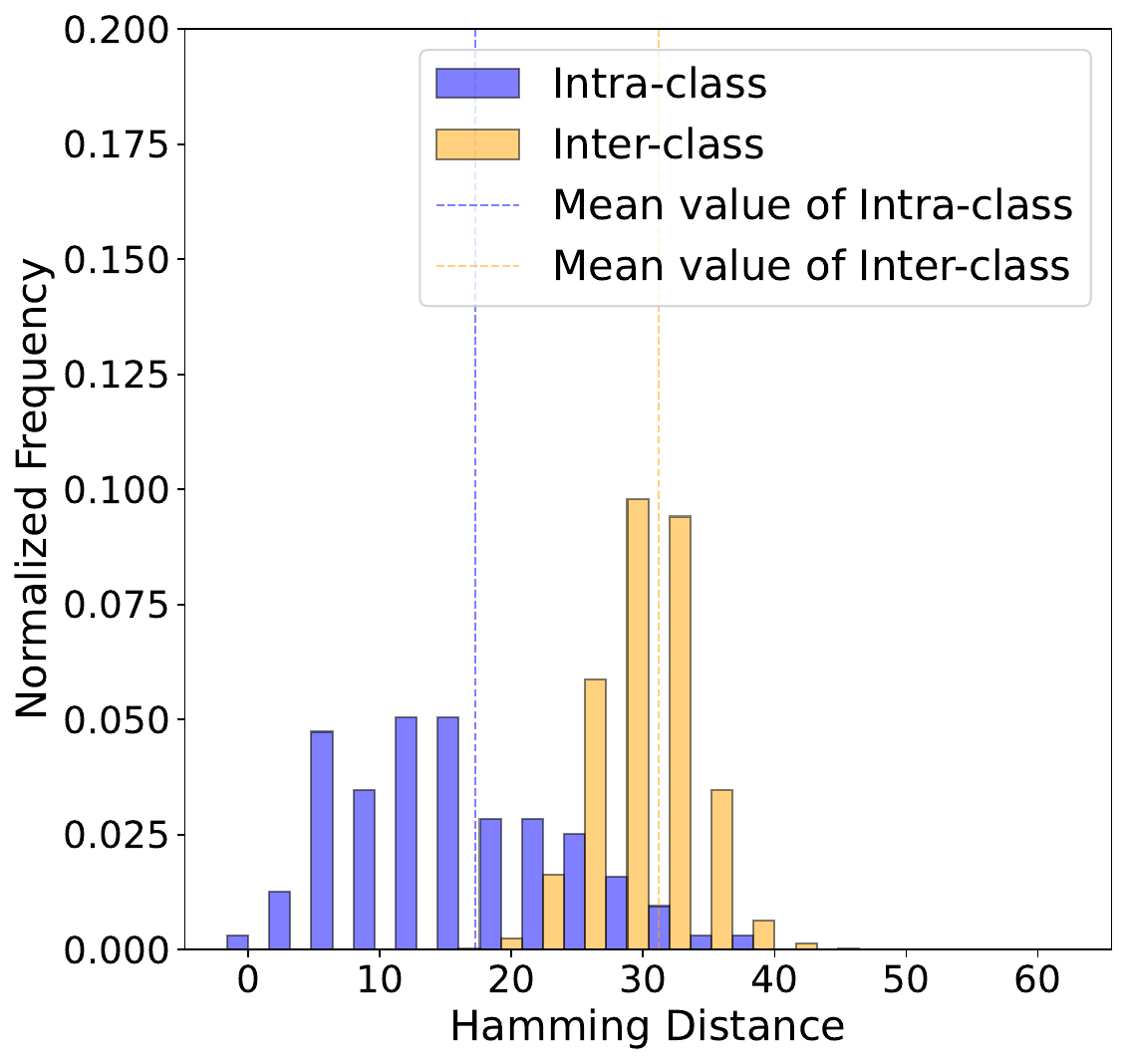}}
    \subfigure[with $\mathcal{L}_\text{classwise}$]{\includegraphics[width=0.4\linewidth]{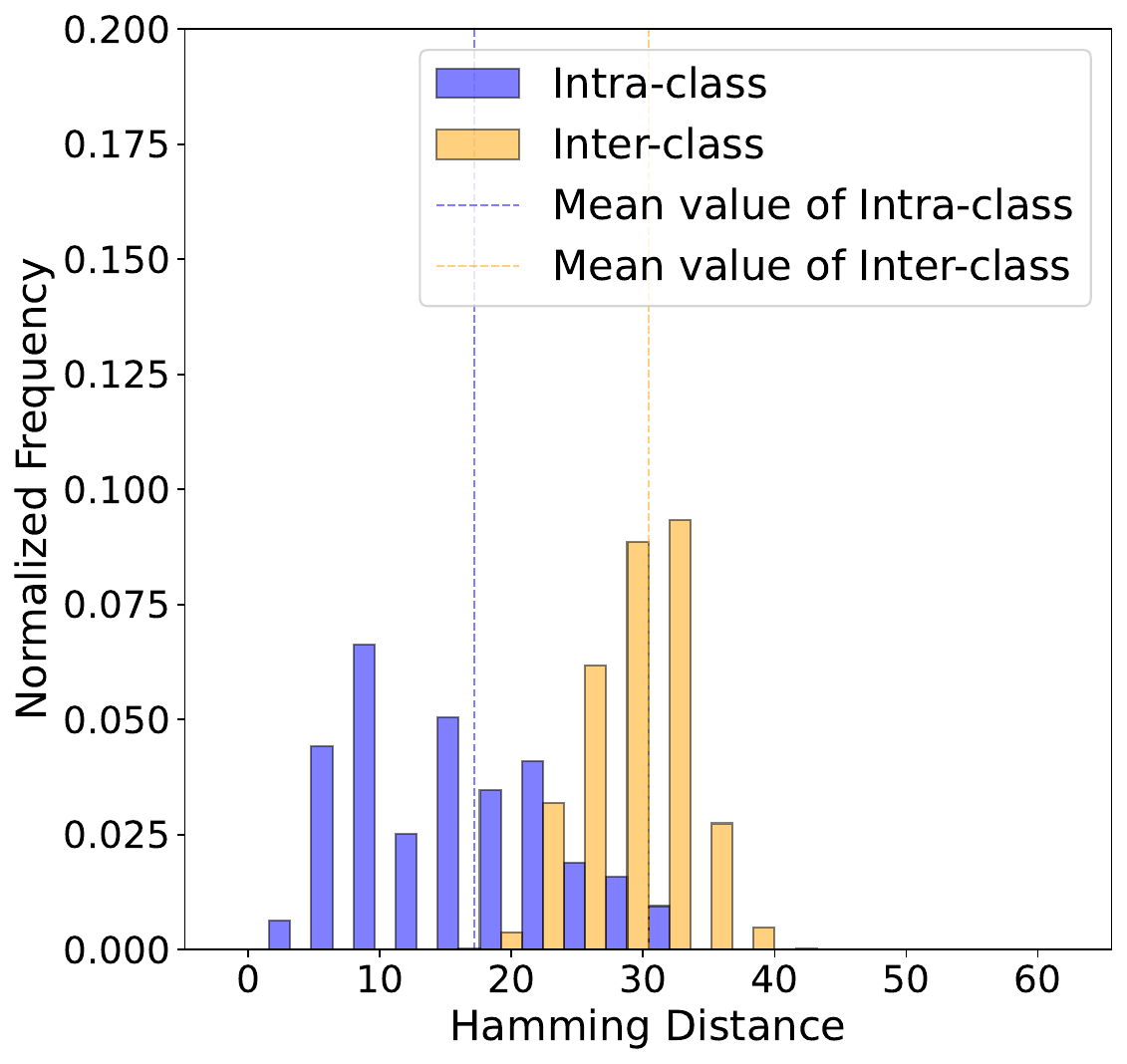}}
    \caption{Histogram distances of the intra-classes and inter-classes. The arrow annotation is the quantitative separability with the hamming distance, $\mathbb{E}[D_{inter}]-\mathbb{E}[D_{intra}]$. }
    % \vspace{-3mm}
    \label{fig:visual_hist_contrastive}
\end{figure}

\subsection{RQ2: Effects of Unseen Class Ratio}

To verify the robustness of our model in the context of the zero-shot scenario, we conduct experiments where the number of seen class categories undergoes a progressive reduction during the training process. The results are visually depicted in Fig. ~\ref{fig:zero_ratio}. Specifically, (d)-(f) and (j)-(l) show the performance decrease ratio between 20\%->40\% and 20\%->60\% respectively.
Based on the outcome results, we conclude the following conclusions: \textbf{(1)} All deep hashing methods show a noticeable decline in performance along with an increase in the number of unseen classes on three datasets. Notably, this performance drop not only intensifies but also amplifies in magnitude, as evidenced by both the MAP@all and AUC metrics. \textbf{(2)} COMAE exhibits a performance superiority across various unseen class ratios, with minimal decrease.

\begin{figure}[!t]
\centering
\includegraphics[width=\linewidth]{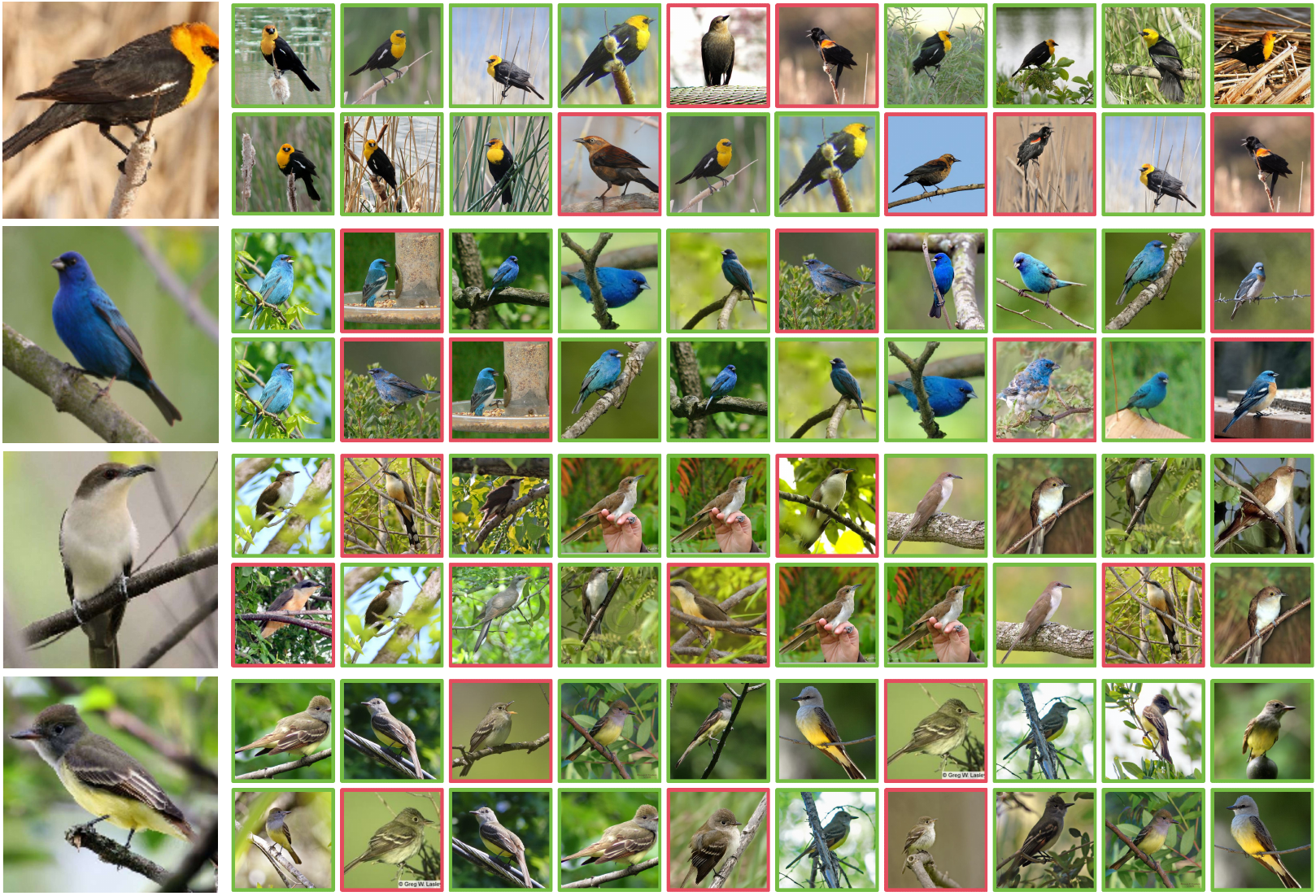}
\caption{Case study on the CUB dataset.} \label{fig:case_study}
% \vspace{-3mm}
\end{figure}

\begin{table}[htbp]
\centering
\caption{The average running time for each epoch}
 \resizebox{\linewidth}{!}{
\label{tab:method_comparison}
\begin{tabular}{c ccc ccc ccc}
\toprule
\multirow{2}{*}{Method}& \multicolumn{3}{c}{AWA2} & \multicolumn{3}{c}{CUB} & \multicolumn{3}{c}{SUN}  \\ \cmidrule(rl){2-4}\cmidrule(rl){5-7}\cmidrule(rl){8-10}
& mAP & Train & Infer. & mAP & Train & Infer. & mAP & Train &Infer.\\ \midrule
PIXEL & 0.5456 & 483s &0.154s & 0.2819 & 625s &0.183s & 0.1741 & 568s &0.231s\\
\textbf{COMAE} & \textbf{0.5801}   & \textbf{45s} & \textbf{0.011s} & \textbf{0.2532} & \textbf{48s } & \textbf{0.052s} & \textbf{0.1764} & \textbf{47s} & \textbf{0.075s}\\ \bottomrule
\end{tabular}
}
% \vspace{-3mm}
\end{table}

\subsection{RQ3: Case Study}
We illustrate the performance of COMAE through case studies with 64-bit hash codes, as shown in Fig~\ref{fig:case_study}. We present the top 10 retrieved images for each query using COMAE (first row) and PIXEL (second row). Correctly retrieved images are outlined in green, while incorrect ones are outlined in red. COMAE demonstrates superior retrieval performance, effectively capturing key visual features. For queries with distinctive visual characteristics, such as the yellow-headed blackbird or the blue bunting, COMAE retrieves highly accurate results, primarily by leveraging attributes like feather color, body shape, and posture.
In more challenging cases involving birds with subtle differences or overlapping features, such as cuckoos and flycatchers, COMAE still maintains strong retrieval accuracy but occasionally retrieves images with semantically similar attributes. For instance, brownish birds with similar body structures sometimes lead to false positives, as observed in some incorrect retrievals.

Overall, for intact retrieval targets, COMAE achieves an accuracy of 80\%, while in complex cases involving partial occlusion, highly similar species, or ambiguous visual features, COMAE retains a 70\% accuracy rate, outperforming PIXEL in zero-shot hashing tasks. The comparison between COMAE and PIXEL highlights COMAE's robustness in learning effective feature representations and reducing semantic drift in image retrieval.

\subsection{RQ4: Efficiency Analysis}
We conduct model efficiency analysis with a 64-bit hashing length in this section. Table \ref{tab:method_comparison} reports the average training time per epoch and inference time per epoch. We select the best and latest baseline, PIXEL, as our comparable baseline. We find that COMAE achieves competitive computational efficiency. In the largest dataset, SUN, COMAE reduces the training time by over 91.7\%, and the inference time is reduced from 0.231s (PIXEL) to 0.075s (COMAE), with an improvement of 67.5\%. This demonstrates that COMAE not only trains significantly faster but also achieves efficient inference, making it well-suited for real world large-scale retrieval systems.

\subsection{RQ5: Model Analysis}
\par \textbf{Distance Visualization of Intra and Inter Classes}. 
For model analysis, we first conduct the distance visualization of intra and inter classes, to verify whether the classwise constraint can broaden the distance of inter-class and reduce the distance of intra-class. We provide a quantitative comparison of the distance between the mean value of intra-class distances and inter-class distances, under the setting of without the contrastive loss $\mathcal{L}_\text{pairwise}$ and with it. 
Fig.~\ref{fig:visual_hist_contrastive}  summarizes the histogram of intra-class and inter-class distances on the AWA2 dataset. We adjust the frequency normalization so that the total of all histogram bins equals one. The directional marker indicates the separability in Hamming distances, $\mathbb{E}[D_{inter}]-\mathbb{E}[D_{intra}]$. Despite similar distributions for inter-class distances, the distinction between the average intra-class distance (indicated by the purple dotted line) and the average inter-class distance (shown by the orange dotted line) in scenario (b) exceeds that in scenario (a). This approach visually emphasizes the effect of the classwise constraint on class separability metrics. Quantificationally, the difference in (a) is  14.38, and the difference in (b) is 15.50. The results show that our proposed classwise constraint can enlarge the distance of the inter-class and reduce the distance of the intra-class.

\noindent{\textbf{Component Analysis.}
To provide further insight into COMAE, we conduct component analysis on AWA2 and CUB datasets with 64-bit hashing-code lengths. We configure several variants of COMAE as (1) \textit{COMAE-V1} is the COMAE model without $\mathcal{L}_\text{pointwise}$; (2) \textit{COMAE-V2} represents the COMAE without $\mathcal{L}_\text{pairwise}$; (3) \textit{COMAE-V3} refers to the COMAE without $\mathcal{L}_\text{classwise}$. Results are shown in Table \ref{tab:component}. 
Compared with other components, the pointwise objective plays the most significant role in AWA2, and the pairwise objective in CUB. Also, the performance of COMAE on both datasets surpasses that of the variants, validating the necessity of each component.}

\begin{table}[htbp]
	\centering
    \caption{The mAP@5000 and AUC results with switching model components in different datasets.}
 \resizebox{\linewidth}{!}{
	\begin{tabular}{cccccccc}
	\toprule
     & \multicolumn{2}{c}{AWA2} & \multicolumn{2}{c}{CUB} & \multicolumn{2}{c}{SUN} \\
     \cmidrule(lr){2-3} \cmidrule(lr){4-5}  \cmidrule(lr){6-7}
     & mAP@5000 & AUC & mAP@5000 & AUC & mAP@5000 & AUC \\
	\midrule
    COMAE-V1 & 0.5093 & 0.4364 & 0.1988 & 0.2333 &0.1420 &0.1286\\
	COMAE-V2 & 0.5122 & 0.4981 & 0.1794 & 0.2161 &0.1310 &0.1243\\
	COMAE-V3  & 0.5099 & 0.5025 & 0.1983 & 0.2327 &0.1350 &0.1264\\\midrule
	\textbf{COMAE}  & \textbf{0.5149} &\textbf{0.5106} & \textbf{0.2012} & \textbf{0.2331} & \textbf{0.1465} & \textbf{0.1302}\\
	\bottomrule
    \end{tabular}}
    \label{tab:component}
    % \vspace{-3mm}
\end{table}

\section{Conclusion}
Zero-Shot Hashing stands out for its high efficiency and robust generalization, particularly in large-scale retrieval scenarios. 
In this paper, we present COMAE for a comprehensive attribute exploration in ZSH. COMAE depicts the relationships between seeable and unseeable classes through  meticulously designed explorations: \textit{point-wise}, \textit{pair-wise}, and \textit{class-wise} constraints. Furthermore, theoretical analysis is provided to show the effectiveness of COMAE. 
Our model employs an attribute prototype network to regress attributes, facilitating the learning of local features. Then, we employ contrastive learning to capture the context of attributes, moving beyond instance-independent optimization. Finally, we introduce the class-wise constraint to cohesively enhance hashing learning, image representations, and visual attributes. Experimental results on the popular ZSH datasets demonstrate that COMAE outperforms state-of-the-art hashing techniques, especially in scenarios with a larger number of unseen label classes.

\begin{acks}
We would like to thank Zeyu Dong, Ning Cao, Shuai Liu, Pengfei Wang, and Yuchen Yan who discussed with us and provided instructive suggestions. This work is partially supported by the National Natural Science Foundation of China (No.92470204), the Beijing Natural Science Foundation (No.4254089), the Postdoctoral Fellowship Program of CPSF (No.GZC20232736), and the China Postdoctoral Science Foundation Funded Project (No.2023M743565).
\end{acks}

\clearpage
% \bibliographystyle{ACM-Reference-Format}
% \bibliography{ref}

\begin{thebibliography}{69}

%%% the Web address.

\ifx \showCODEN    \undefined \def \showCODEN     #1{\unskip}     \fi
\ifx \showISBNx    \undefined \def \showISBNx     #1{\unskip}     \fi
\ifx \showISBNxiii \undefined \def \showISBNxiii  #1{\unskip}     \fi
\ifx \showISSN     \undefined \def \showISSN      #1{\unskip}     \fi
\ifx \showLCCN     \undefined \def \showLCCN      #1{\unskip}     \fi
\ifx \shownote     \undefined \def \shownote      #1{#1}          \fi
\ifx \showarticletitle \undefined \def \showarticletitle #1{#1}   \fi
\ifx \showURL      \undefined \def \showURL       {\relax}        \fi
% invisible to TeX
\providecommand\bibfield[2]{#2}
\providecommand\bibinfo[2]{#2}
\providecommand\natexlab[1]{#1}
\providecommand\showeprint[2][]{arXiv:#2}

\bibitem[Andoni and Indyk(2008)]%
        {andoni2008near}
\bibfield{author}{\bibinfo{person}{Alexandr Andoni} {and} \bibinfo{person}{Piotr Indyk}.} \bibinfo{year}{2008}\natexlab{}.
\newblock \showarticletitle{Near-optimal hashing algorithms for approximate nearest neighbor in high dimensions}.
\newblock \bibinfo{journal}{\emph{Commun. ACM}} \bibinfo{volume}{51}, \bibinfo{number}{1} (\bibinfo{year}{2008}), \bibinfo{pages}{117--122}.
\newblock


\bibitem[Arik and Pfister(2019)]%
        {arik2019attention}
\bibfield{author}{\bibinfo{person}{Sercan~{\"O}mer Arik} {and} \bibinfo{person}{Tomas Pfister}.} \bibinfo{year}{2019}\natexlab{}.
\newblock \showarticletitle{Attention-Based Prototypical Learning Towards Interpretable, Confident and Robust Deep Neural Networks}.
\newblock \bibinfo{journal}{\emph{arXiv preprint arXiv:1902.06292}} (\bibinfo{year}{2019}).
\newblock


\bibitem[Cai et~al\mbox{.}(2023)]%
        {cai2023resolving}
\bibfield{author}{\bibinfo{person}{Xunxin Cai}, \bibinfo{person}{Meng Xiao}, \bibinfo{person}{Zhiyuan Ning}, {and} \bibinfo{person}{Yuanchun Zhou}.} \bibinfo{year}{2023}\natexlab{}.
\newblock \showarticletitle{Resolving the Imbalance Issue in Hierarchical Disciplinary Topic Inference via LLM-based Data Augmentation}. In \bibinfo{booktitle}{\emph{2023 IEEE International Conference on Data Mining (ICDM)}}. IEEE, \bibinfo{pages}{956--961}.
\newblock


\bibitem[Cao et~al\mbox{.}(2017)]%
        {cao2017hashnet}
\bibfield{author}{\bibinfo{person}{Zhangjie Cao}, \bibinfo{person}{Mingsheng Long}, \bibinfo{person}{Jianmin Wang}, {and} \bibinfo{person}{Philip~S Yu}.} \bibinfo{year}{2017}\natexlab{}.
\newblock \showarticletitle{Hashnet: Deep learning to hash by continuation}. In \bibinfo{booktitle}{\emph{Proceedings of the IEEE international conference on computer vision}}. \bibinfo{pages}{5608--5617}.
\newblock


\bibitem[Chen et~al\mbox{.}(2025b)]%
        {chen2025ambiguity}
\bibfield{author}{\bibinfo{person}{Junyu Chen}, \bibinfo{person}{Yihua Gao}, \bibinfo{person}{Mingyuan Ge}, {and} \bibinfo{person}{Mingyong Li}.} \bibinfo{year}{2025}\natexlab{b}.
\newblock \showarticletitle{Ambiguity-Aware and High-order Relation learning for multi-grained image--text matching}.
\newblock \bibinfo{journal}{\emph{Knowledge-Based Systems}} (\bibinfo{year}{2025}).
\newblock


\bibitem[Chen et~al\mbox{.}(2025a)]%
        {chen2025VSD}
\bibfield{author}{\bibinfo{person}{Junyu Chen}, \bibinfo{person}{Yihua Gao}, {and} \bibinfo{person}{Mingyong Li}.} \bibinfo{year}{2025}\natexlab{a}.
\newblock \showarticletitle{Visual Semantic Description Generation with MLLMs for Image-Text Matching}. In \bibinfo{booktitle}{\emph{ICME}}. IEEE.
\newblock


\bibitem[Chen et~al\mbox{.}(2022)]%
        {chen2022transzero}
\bibfield{author}{\bibinfo{person}{Shiming Chen}, \bibinfo{person}{Ziming Hong}, \bibinfo{person}{Yang Liu}, \bibinfo{person}{Guo-Sen Xie}, \bibinfo{person}{Baigui Sun}, \bibinfo{person}{Hao Li}, \bibinfo{person}{Qinmu Peng}, \bibinfo{person}{Ke Lu}, {and} \bibinfo{person}{Xinge You}.} \bibinfo{year}{2022}\natexlab{}.
\newblock \showarticletitle{Transzero: Attribute-guided transformer for zero-shot learning}. In \bibinfo{booktitle}{\emph{Proceedings of the AAAI Conference on Artificial Intelligence}}, Vol.~\bibinfo{volume}{36}. \bibinfo{pages}{330--338}.
\newblock


\bibitem[Chen et~al\mbox{.}(2018)]%
        {chen2018contrastive}
\bibfield{author}{\bibinfo{person}{Yi Chen}, \bibinfo{person}{Zhuoran Yang}, \bibinfo{person}{Yuchen Xie}, {and} \bibinfo{person}{Zhaoran Wang}.} \bibinfo{year}{2018}\natexlab{}.
\newblock \showarticletitle{Contrastive learning from pairwise measurements}.
\newblock \bibinfo{journal}{\emph{Advances in Neural Information Processing Systems}}  \bibinfo{volume}{31} (\bibinfo{year}{2018}).
\newblock


\bibitem[Chu et~al\mbox{.}(2025)]%
        {chu2025cross}
\bibfield{author}{\bibinfo{person}{Jiaming Chu}, \bibinfo{person}{Yanzhuo Xiang}, \bibinfo{person}{Yuqi Li}, \bibinfo{person}{Chuanguang Yang}, \bibinfo{person}{Zhulin An}, {and} \bibinfo{person}{Yongjun Xu}.} \bibinfo{year}{2025}\natexlab{}.
\newblock \showarticletitle{Cross-Layer Graph Knowledge Distillation for Image Recognition}. In \bibinfo{booktitle}{\emph{ICASSP}}.
\newblock


\bibitem[Dong et~al\mbox{.}(2024a)]%
        {dong2024pixel}
\bibfield{author}{\bibinfo{person}{Zeyu Dong}, \bibinfo{person}{Qingqing Long}, \bibinfo{person}{Yihang Zhou}, \bibinfo{person}{Pengfei Wang}, \bibinfo{person}{Zhihong Zhu}, \bibinfo{person}{Xiao Luo}, \bibinfo{person}{Yidong Wang}, \bibinfo{person}{Pengyang Wang}, {and} \bibinfo{person}{Yuanchun Zhou}.} \bibinfo{year}{2024}\natexlab{a}.
\newblock \showarticletitle{PIXEL: Prompt-based Zero-shot Hashing via Visual and Textual Semantic Alignment}. In \bibinfo{booktitle}{\emph{Proceedings of the 33rd ACM International Conference on Information and Knowledge Management}}. \bibinfo{pages}{487--496}.
\newblock


\bibitem[Dong et~al\mbox{.}(2024b)]%
        {dong2024class}
\bibfield{author}{\bibinfo{person}{Zeyu Dong}, \bibinfo{person}{Chuanguang Yang}, \bibinfo{person}{Yuqi Li}, \bibinfo{person}{Libo Huang}, \bibinfo{person}{Zhulin An}, {and} \bibinfo{person}{Yongjun Xu}.} \bibinfo{year}{2024}\natexlab{b}.
\newblock \showarticletitle{Class-wise Image Mixture Guided Self-Knowledge Distillation for Image Classification}. In \bibinfo{booktitle}{\emph{CSCWD}}. IEEE.
\newblock


\bibitem[Fan et~al\mbox{.}(2020)]%
        {fan2020deep}
\bibfield{author}{\bibinfo{person}{Lixin Fan}, \bibinfo{person}{Kam~Woh Ng}, \bibinfo{person}{Ce Ju}, \bibinfo{person}{Tianyu Zhang}, {and} \bibinfo{person}{Chee~Seng Chan}.} \bibinfo{year}{2020}\natexlab{}.
\newblock \showarticletitle{Deep Polarized Network for Supervised Learning of Accurate Binary Hashing Codes.}. In \bibinfo{booktitle}{\emph{IJCAI}}. \bibinfo{pages}{825--831}.
\newblock


\bibitem[Farhadi et~al\mbox{.}(2009)]%
        {farhadi2009describing}
\bibfield{author}{\bibinfo{person}{Ali Farhadi}, \bibinfo{person}{Ian Endres}, \bibinfo{person}{Derek Hoiem}, {and} \bibinfo{person}{David Forsyth}.} \bibinfo{year}{2009}\natexlab{}.
\newblock \showarticletitle{Describing objects by their attributes}. In \bibinfo{booktitle}{\emph{2009 IEEE conference on computer vision and pattern recognition}}. IEEE, \bibinfo{pages}{1778--1785}.
\newblock


\bibitem[Gao et~al\mbox{.}(2019)]%
        {gao2019hybrid}
\bibfield{author}{\bibinfo{person}{Tianyu Gao}, \bibinfo{person}{Xu Han}, \bibinfo{person}{Zhiyuan Liu}, {and} \bibinfo{person}{Maosong Sun}.} \bibinfo{year}{2019}\natexlab{}.
\newblock \showarticletitle{Hybrid attention-based prototypical networks for noisy few-shot relation classification}. In \bibinfo{booktitle}{\emph{Proceedings of the AAAI conference on artificial intelligence}}, Vol.~\bibinfo{volume}{33}. \bibinfo{pages}{6407--6414}.
\newblock


\bibitem[Gong et~al\mbox{.}(2012)]%
        {gong2012iterative}
\bibfield{author}{\bibinfo{person}{Yunchao Gong}, \bibinfo{person}{Svetlana Lazebnik}, \bibinfo{person}{Albert Gordo}, {and} \bibinfo{person}{Florent Perronnin}.} \bibinfo{year}{2012}\natexlab{}.
\newblock \showarticletitle{Iterative quantization: A procrustean approach to learning binary codes for large-scale image retrieval}.
\newblock \bibinfo{journal}{\emph{IEEE transactions on pattern analysis and machine intelligence}} \bibinfo{volume}{35}, \bibinfo{number}{12} (\bibinfo{year}{2012}), \bibinfo{pages}{2916--2929}.
\newblock


\bibitem[Guo et~al\mbox{.}(2025)]%
        {guo2025deepseek}
\bibfield{author}{\bibinfo{person}{Daya Guo}, \bibinfo{person}{Dejian Yang}, \bibinfo{person}{Haowei Zhang}, \bibinfo{person}{Junxiao Song}, \bibinfo{person}{Ruoyu Zhang}, \bibinfo{person}{Runxin Xu}, \bibinfo{person}{Qihao Zhu}, \bibinfo{person}{Shirong Ma}, \bibinfo{person}{Peiyi Wang}, \bibinfo{person}{Xiao Bi}, {et~al\mbox{.}}} \bibinfo{year}{2025}\natexlab{}.
\newblock \showarticletitle{Deepseek-r1: Incentivizing reasoning capability in llms via reinforcement learning}.
\newblock \bibinfo{journal}{\emph{arXiv preprint arXiv:2501.12948}} (\bibinfo{year}{2025}).
\newblock


\bibitem[Guo et~al\mbox{.}(2017)]%
        {guo2017sitnet}
\bibfield{author}{\bibinfo{person}{Yuchen Guo}, \bibinfo{person}{Guiguang Ding}, \bibinfo{person}{Jungong Han}, {and} \bibinfo{person}{Yue Gao}.} \bibinfo{year}{2017}\natexlab{}.
\newblock \showarticletitle{SitNet: Discrete Similarity Transfer Network for Zero-shot Hashing.}. In \bibinfo{booktitle}{\emph{IJCAI}}. \bibinfo{pages}{1767--1773}.
\newblock


\bibitem[Han et~al\mbox{.}(2021)]%
        {han2021contrastive}
\bibfield{author}{\bibinfo{person}{Zongyan Han}, \bibinfo{person}{Zhenyong Fu}, \bibinfo{person}{Shuo Chen}, {and} \bibinfo{person}{Jian Yang}.} \bibinfo{year}{2021}\natexlab{}.
\newblock \showarticletitle{Contrastive embedding for generalized zero-shot learning}. In \bibinfo{booktitle}{\emph{Proceedings of the IEEE/CVF conference on computer vision and pattern recognition}}. \bibinfo{pages}{2371--2381}.
\newblock


\bibitem[Hoe et~al\mbox{.}(2021)]%
        {hoe2021one}
\bibfield{author}{\bibinfo{person}{Jiun~Tian Hoe}, \bibinfo{person}{Kam~Woh Ng}, \bibinfo{person}{Tianyu Zhang}, \bibinfo{person}{Chee~Seng Chan}, \bibinfo{person}{Yi-Zhe Song}, {and} \bibinfo{person}{Tao Xiang}.} \bibinfo{year}{2021}\natexlab{}.
\newblock \showarticletitle{One loss for all: Deep hashing with a single cosine similarity based learning objective}.
\newblock \bibinfo{journal}{\emph{Advances in Neural Information Processing Systems}}  \bibinfo{volume}{34} (\bibinfo{year}{2021}), \bibinfo{pages}{24286--24298}.
\newblock


\bibitem[Khan et~al\mbox{.}(2018)]%
        {khan2018evaluating}
\bibfield{author}{\bibinfo{person}{Riaz~Ullah Khan}, \bibinfo{person}{Xiaosong Zhang}, \bibinfo{person}{Rajesh Kumar}, {and} \bibinfo{person}{Emelia~Opoku Aboagye}.} \bibinfo{year}{2018}\natexlab{}.
\newblock \showarticletitle{Evaluating the performance of resnet model based on image recognition}. In \bibinfo{booktitle}{\emph{Proceedings of the 2018 International Conference on Computing and Artificial Intelligence}}. \bibinfo{pages}{86--90}.
\newblock


\bibitem[Li et~al\mbox{.}(2025c)]%
        {li2025wavfusion}
\bibfield{author}{\bibinfo{person}{Feng Li}, \bibinfo{person}{Jiusong Luo}, {and} \bibinfo{person}{Wanjun Xia}.} \bibinfo{year}{2025}\natexlab{c}.
\newblock \showarticletitle{WavFusion: Towards Wav2vec 2.0 Multimodal Speech Emotion Recognition}. In \bibinfo{booktitle}{\emph{International Conference on Multimedia Modeling}}. Springer, \bibinfo{pages}{325--336}.
\newblock


\bibitem[Li et~al\mbox{.}(2021b)]%
        {li2021explicit}
\bibfield{author}{\bibinfo{person}{Feng Li}, \bibinfo{person}{Bencheng Yan}, \bibinfo{person}{Qingqing Long}, \bibinfo{person}{Pengjie Wang}, \bibinfo{person}{Wei Lin}, \bibinfo{person}{Jian Xu}, {and} \bibinfo{person}{Bo Zheng}.} \bibinfo{year}{2021}\natexlab{b}.
\newblock \showarticletitle{Explicit semantic cross feature learning via pre-trained graph neural networks for CTR prediction}. In \bibinfo{booktitle}{\emph{Proceedings of the 44th international ACM SIGIR conference on research and development in information retrieval}}. \bibinfo{pages}{2161--2165}.
\newblock


\bibitem[Li et~al\mbox{.}(2021a)]%
        {li2021adaptive}
\bibfield{author}{\bibinfo{person}{Gen Li}, \bibinfo{person}{Varun Jampani}, \bibinfo{person}{Laura Sevilla-Lara}, \bibinfo{person}{Deqing Sun}, \bibinfo{person}{Jonghyun Kim}, {and} \bibinfo{person}{Joongkyu Kim}.} \bibinfo{year}{2021}\natexlab{a}.
\newblock \showarticletitle{Adaptive prototype learning and allocation for few-shot segmentation}. In \bibinfo{booktitle}{\emph{Proceedings of the IEEE/CVF conference on computer vision and pattern recognition}}. \bibinfo{pages}{8334--8343}.
\newblock


\bibitem[Li et~al\mbox{.}(2025d)]%
        {li2025prototype}
\bibfield{author}{\bibinfo{person}{Jiarui Li}, \bibinfo{person}{Qiu Zhen}, \bibinfo{person}{Yilin Yang}, \bibinfo{person}{Yuqi Li}, \bibinfo{person}{Zeyu Dong}, {and} \bibinfo{person}{Chuanguang Yang}.} \bibinfo{year}{2025}\natexlab{d}.
\newblock \showarticletitle{Prototype-Driven Multi-Feature Generation for Visible-Infrared Person Re-identification}. In \bibinfo{booktitle}{\emph{ICASSP2025}}. IEEE.
\newblock


\bibitem[Li et~al\mbox{.}(2025a)]%
        {li2025PASE}
\bibfield{author}{\bibinfo{person}{Mingyong Li}, \bibinfo{person}{Yihua Gao}, \bibinfo{person}{Honggang Zhao}, \bibinfo{person}{Ruiheng Li}, {and} \bibinfo{person}{Junyu Chen}.} \bibinfo{year}{2025}\natexlab{a}.
\newblock \showarticletitle{Progressive semantic aggregation and structured cognitive enhancement for image--text matching}.
\newblock \bibinfo{journal}{\emph{Expert Systems with Applications}}  \bibinfo{volume}{274} (\bibinfo{year}{2025}), \bibinfo{pages}{126943}.
\newblock


\bibitem[Li et~al\mbox{.}(2025b)]%
        {li2025fedkd}
\bibfield{author}{\bibinfo{person}{Yuqi Li}, \bibinfo{person}{Xingyou Lin}, \bibinfo{person}{Kai Zhang}, \bibinfo{person}{Chuanguang Yang}, \bibinfo{person}{Zhongliang Guo}, \bibinfo{person}{Jianping Gou}, {and} \bibinfo{person}{Yanli Li}.} \bibinfo{year}{2025}\natexlab{b}.
\newblock \showarticletitle{FedKD-hybrid: Federated Hybrid Knowledge Distillation for Lithography Hotspot Detection}.
\newblock \bibinfo{journal}{\emph{arXiv preprint arXiv:2501.04066}} (\bibinfo{year}{2025}).
\newblock


\bibitem[Li et~al\mbox{.}(2024)]%
        {li2024sglp}
\bibfield{author}{\bibinfo{person}{Yuqi Li}, \bibinfo{person}{Yao Lu}, \bibinfo{person}{Zeyu Dong}, \bibinfo{person}{Chuanguang Yang}, \bibinfo{person}{Yihao Chen}, {and} \bibinfo{person}{Jianping Gou}.} \bibinfo{year}{2024}\natexlab{}.
\newblock \showarticletitle{SGLP: A Similarity Guided Fast Layer Partition Pruning for Compressing Large Deep Models}.
\newblock \bibinfo{journal}{\emph{arXiv preprint arXiv:2410.14720}} (\bibinfo{year}{2024}).
\newblock


\bibitem[Li and van Gemert(2021)]%
        {li2021deep}
\bibfield{author}{\bibinfo{person}{Yunqiang Li} {and} \bibinfo{person}{Jan van Gemert}.} \bibinfo{year}{2021}\natexlab{}.
\newblock \showarticletitle{Deep unsupervised image hashing by maximizing bit entropy}. In \bibinfo{booktitle}{\emph{Proceedings of the AAAI Conference on Artificial Intelligence}}, Vol.~\bibinfo{volume}{35}. \bibinfo{pages}{2002--2010}.
\newblock


\bibitem[Lin et~al\mbox{.}(2024)]%
        {lin2024gume}
\bibfield{author}{\bibinfo{person}{Guojiao Lin}, \bibinfo{person}{Meng Zhen}, \bibinfo{person}{Dongjie Wang}, \bibinfo{person}{Qingqing Long}, \bibinfo{person}{Yuanchun Zhou}, {and} \bibinfo{person}{Meng Xiao}.} \bibinfo{year}{2024}\natexlab{}.
\newblock \showarticletitle{GUME: Graphs and User Modalities Enhancement for Long-Tail Multimodal Recommendation}. In \bibinfo{booktitle}{\emph{Proceedings of the 33rd ACM International Conference on Information and Knowledge Management}}. \bibinfo{pages}{1400--1409}.
\newblock


\bibitem[Liu et~al\mbox{.}(2017)]%
        {liu2017sphereface}
\bibfield{author}{\bibinfo{person}{Weiyang Liu}, \bibinfo{person}{Yandong Wen}, \bibinfo{person}{Zhiding Yu}, \bibinfo{person}{Ming Li}, \bibinfo{person}{Bhiksha Raj}, {and} \bibinfo{person}{Le Song}.} \bibinfo{year}{2017}\natexlab{}.
\newblock \showarticletitle{Sphereface: Deep hypersphere embedding for face recognition}. In \bibinfo{booktitle}{\emph{Proceedings of the IEEE conference on computer vision and pattern recognition}}. \bibinfo{pages}{212--220}.
\newblock


\bibitem[Long et~al\mbox{.}(2020)]%
        {long2020graph}
\bibfield{author}{\bibinfo{person}{Qingqing Long}, \bibinfo{person}{Yilun Jin}, \bibinfo{person}{Guojie Song}, \bibinfo{person}{Yi Li}, {and} \bibinfo{person}{Wei Lin}.} \bibinfo{year}{2020}\natexlab{}.
\newblock \showarticletitle{Graph structural-topic neural network}. In \bibinfo{booktitle}{\emph{Proceedings of the 26th ACM SIGKDD international conference on knowledge discovery \& data mining}}. \bibinfo{pages}{1065--1073}.
\newblock


\bibitem[Long et~al\mbox{.}(2021)]%
        {long2021hgk}
\bibfield{author}{\bibinfo{person}{Qingqing Long}, \bibinfo{person}{Lingjun Xu}, \bibinfo{person}{Zheng Fang}, {and} \bibinfo{person}{Guojie Song}.} \bibinfo{year}{2021}\natexlab{}.
\newblock \showarticletitle{Hgk-gnn: Heterogeneous graph kernel based graph neural networks}. In \bibinfo{booktitle}{\emph{Proceedings of the 27th ACM SIGKDD conference on knowledge discovery \& data mining}}. \bibinfo{pages}{1129--1138}.
\newblock


\bibitem[Lu et~al\mbox{.}(2024)]%
        {lu2024generic}
\bibfield{author}{\bibinfo{person}{Yao Lu}, \bibinfo{person}{Yutao Zhu}, \bibinfo{person}{Yuqi Li}, \bibinfo{person}{Dongwei Xu}, \bibinfo{person}{Yun Lin}, \bibinfo{person}{Qi Xuan}, {and} \bibinfo{person}{Xiaoniu Yang}.} \bibinfo{year}{2024}\natexlab{}.
\newblock \showarticletitle{A generic layer pruning method for signal modulation recognition deep learning models}.
\newblock \bibinfo{journal}{\emph{IEEE TCCN}} (\bibinfo{year}{2024}).
\newblock


\bibitem[Luo et~al\mbox{.}(2025)]%
        {luo2025large}
\bibfield{author}{\bibinfo{person}{Junyu Luo}, \bibinfo{person}{Weizhi Zhang}, \bibinfo{person}{Ye Yuan}, \bibinfo{person}{Yusheng Zhao}, \bibinfo{person}{Junwei Yang}, \bibinfo{person}{Yiyang Gu}, \bibinfo{person}{Bohan Wu}, \bibinfo{person}{Binqi Chen}, \bibinfo{person}{Ziyue Qiao}, {et~al\mbox{.}}} \bibinfo{year}{2025}\natexlab{}.
\newblock \showarticletitle{Large Language Model Agent: A Survey on Methodology, Applications and Challenges}.
\newblock \bibinfo{journal}{\emph{arXiv preprint arXiv:2503.21460}} (\bibinfo{year}{2025}).
\newblock


\bibitem[Ma et~al\mbox{.}(2024)]%
        {ma2024discrepancy}
\bibfield{author}{\bibinfo{person}{Zeyu Ma}, \bibinfo{person}{Yuqi Li}, \bibinfo{person}{Yizhi Luo}, \bibinfo{person}{Xiao Luo}, \bibinfo{person}{Jinxing Li}, \bibinfo{person}{Chong Chen}, \bibinfo{person}{Xian-Sheng Hua}, {and} \bibinfo{person}{Guangming Lu}.} \bibinfo{year}{2024}\natexlab{}.
\newblock \showarticletitle{Discrepancy and structure-based contrast for test-time adaptive retrieval}.
\newblock \bibinfo{journal}{\emph{IEEE Transactions on Multimedia}} (\bibinfo{year}{2024}).
\newblock


\bibitem[Ma{\'c}kiewicz and Ratajczak(1993)]%
        {mackiewicz1993principal}
\bibfield{author}{\bibinfo{person}{Andrzej Ma{\'c}kiewicz} {and} \bibinfo{person}{Waldemar Ratajczak}.} \bibinfo{year}{1993}\natexlab{}.
\newblock \showarticletitle{Principal components analysis (PCA)}.
\newblock \bibinfo{journal}{\emph{Computers \& Geosciences}} \bibinfo{volume}{19}, \bibinfo{number}{3} (\bibinfo{year}{1993}), \bibinfo{pages}{303--342}.
\newblock


\bibitem[Patterson and Hays(2012)]%
        {patterson2012sun}
\bibfield{author}{\bibinfo{person}{Genevieve Patterson} {and} \bibinfo{person}{James Hays}.} \bibinfo{year}{2012}\natexlab{}.
\newblock \showarticletitle{Sun attribute database: Discovering, annotating, and recognizing scene attributes}. In \bibinfo{booktitle}{\emph{2012 IEEE conference on computer vision and pattern recognition}}. IEEE, \bibinfo{pages}{2751--2758}.
\newblock


\bibitem[Pourpanah et~al\mbox{.}(2022)]%
        {pourpanah2022review}
\bibfield{author}{\bibinfo{person}{Farhad Pourpanah}, \bibinfo{person}{Moloud Abdar}, \bibinfo{person}{Yuxuan Luo}, \bibinfo{person}{Xinlei Zhou}, \bibinfo{person}{Ran Wang}, \bibinfo{person}{Chee~Peng Lim}, \bibinfo{person}{Xi-Zhao Wang}, {and} \bibinfo{person}{QM~Jonathan Wu}.} \bibinfo{year}{2022}\natexlab{}.
\newblock \showarticletitle{A review of generalized zero-shot learning methods}.
\newblock \bibinfo{journal}{\emph{IEEE transactions on pattern analysis and machine intelligence}} (\bibinfo{year}{2022}).
\newblock


\bibitem[Qin et~al\mbox{.}(2023)]%
        {qin2023destruction}
\bibfield{author}{\bibinfo{person}{Tianrui Qin}, \bibinfo{person}{Xitong Gao}, \bibinfo{person}{Juanjuan Zhao}, {and} \bibinfo{person}{Kejiang Ye}.} \bibinfo{year}{2023}\natexlab{}.
\newblock \showarticletitle{Destruction-Restoration Suppresses Data Protection Perturbations against Diffusion Models}. In \bibinfo{booktitle}{\emph{ICTAI}}. IEEE.
\newblock


\bibitem[Qin et~al\mbox{.}(2024)]%
        {qinapbench}
\bibfield{author}{\bibinfo{person}{Tianrui Qin}, \bibinfo{person}{Xitong Gao}, \bibinfo{person}{Juanjuan Zhao}, \bibinfo{person}{Kejiang Ye}, {and} \bibinfo{person}{Cheng-zhong Xu}.} \bibinfo{year}{2024}\natexlab{}.
\newblock \showarticletitle{APBench: A Unified Availability Poisoning Attack and Defenses Benchmark}.
\newblock \bibinfo{journal}{\emph{Transactions on Machine Learning Research}} (\bibinfo{year}{2024}).
\newblock


\bibitem[Qin et~al\mbox{.}(2022)]%
        {qin2022flareon}
\bibfield{author}{\bibinfo{person}{Tianrui Qin}, \bibinfo{person}{Xianghuan He}, \bibinfo{person}{Xitong Gao}, \bibinfo{person}{Yiren Zhao}, \bibinfo{person}{Kejiang Ye}, {and} \bibinfo{person}{Cheng-Zhong Xu}.} \bibinfo{year}{2022}\natexlab{}.
\newblock \showarticletitle{Flareon: Stealthy any2any backdoor injection via poisoned augmentation}.
\newblock \bibinfo{journal}{\emph{arXiv preprint arXiv:2212.09979}} (\bibinfo{year}{2022}).
\newblock


\bibitem[Qiu et~al\mbox{.}(2021)]%
        {qiu2021unsupervised}
\bibfield{author}{\bibinfo{person}{Zexuan Qiu}, \bibinfo{person}{Qinliang Su}, \bibinfo{person}{Zijing Ou}, \bibinfo{person}{Jianxing Yu}, {and} \bibinfo{person}{Changyou Chen}.} \bibinfo{year}{2021}\natexlab{}.
\newblock \showarticletitle{Unsupervised hashing with contrastive information bottleneck}.
\newblock \bibinfo{journal}{\emph{IJCAI}} (\bibinfo{year}{2021}).
\newblock


\bibitem[Shen et~al\mbox{.}(2015)]%
        {shen2015hashing}
\bibfield{author}{\bibinfo{person}{Fumin Shen}, \bibinfo{person}{Chunhua Shen}, \bibinfo{person}{Qinfeng Shi}, \bibinfo{person}{Anton Van~den Hengel}, \bibinfo{person}{Zhenmin Tang}, {and} \bibinfo{person}{Heng~Tao Shen}.} \bibinfo{year}{2015}\natexlab{}.
\newblock \showarticletitle{Hashing on nonlinear manifolds}.
\newblock \bibinfo{journal}{\emph{IEEE Transactions on Image Processing}} \bibinfo{volume}{24}, \bibinfo{number}{6} (\bibinfo{year}{2015}), \bibinfo{pages}{1839--1851}.
\newblock


\bibitem[Shen et~al\mbox{.}(2019)]%
        {shen2019embarrassingly}
\bibfield{author}{\bibinfo{person}{Yuming Shen}, \bibinfo{person}{Jie Qin}, \bibinfo{person}{Jiaxin Chen}, \bibinfo{person}{Li Liu}, \bibinfo{person}{Fan Zhu}, {and} \bibinfo{person}{Ziyi Shen}.} \bibinfo{year}{2019}\natexlab{}.
\newblock \showarticletitle{Embarrassingly simple binary representation learning}. In \bibinfo{booktitle}{\emph{Proceedings of the IEEE/CVF International Conference on Computer Vision Workshops}}. \bibinfo{pages}{0--0}.
\newblock


\bibitem[Shen et~al\mbox{.}(2020)]%
        {shen2020auto}
\bibfield{author}{\bibinfo{person}{Yuming Shen}, \bibinfo{person}{Jie Qin}, \bibinfo{person}{Jiaxin Chen}, \bibinfo{person}{Mengyang Yu}, \bibinfo{person}{Li Liu}, \bibinfo{person}{Fan Zhu}, \bibinfo{person}{Fumin Shen}, {and} \bibinfo{person}{Ling Shao}.} \bibinfo{year}{2020}\natexlab{}.
\newblock \showarticletitle{Auto-encoding twin-bottleneck hashing}. In \bibinfo{booktitle}{\emph{Proceedings of the IEEE/CVF conference on computer vision and pattern recognition}}. \bibinfo{pages}{2818--2827}.
\newblock


\bibitem[Shi et~al\mbox{.}(2022)]%
        {shi2022zero}
\bibfield{author}{\bibinfo{person}{Yang Shi}, \bibinfo{person}{Xiushan Nie}, \bibinfo{person}{Xingbo Liu}, \bibinfo{person}{Lu Yang}, {and} \bibinfo{person}{Yilong Yin}.} \bibinfo{year}{2022}\natexlab{}.
\newblock \showarticletitle{Zero-shot hashing via asymmetric ratio similarity matrix}.
\newblock \bibinfo{journal}{\emph{IEEE Transactions on Knowledge and Data Engineering}} \bibinfo{volume}{35}, \bibinfo{number}{5} (\bibinfo{year}{2022}), \bibinfo{pages}{5426--5437}.
\newblock


\bibitem[Su et~al\mbox{.}(2018)]%
        {su2018greedy}
\bibfield{author}{\bibinfo{person}{Shupeng Su}, \bibinfo{person}{Chao Zhang}, \bibinfo{person}{Kai Han}, {and} \bibinfo{person}{Yonghong Tian}.} \bibinfo{year}{2018}\natexlab{}.
\newblock \showarticletitle{Greedy hash: Towards fast optimization for accurate hash coding in cnn}.
\newblock \bibinfo{journal}{\emph{Advances in neural information processing systems}}  \bibinfo{volume}{31} (\bibinfo{year}{2018}).
\newblock


\bibitem[Sylvain et~al\mbox{.}(2019)]%
        {sylvain2019locality}
\bibfield{author}{\bibinfo{person}{Tristan Sylvain}, \bibinfo{person}{Linda Petrini}, {and} \bibinfo{person}{Devon Hjelm}.} \bibinfo{year}{2019}\natexlab{}.
\newblock \showarticletitle{Locality and compositionality in zero-shot learning}.
\newblock \bibinfo{journal}{\emph{arXiv preprint arXiv:1912.12179}} (\bibinfo{year}{2019}).
\newblock


\bibitem[Tan et~al\mbox{.}(2022)]%
        {tan2022fedproto}
\bibfield{author}{\bibinfo{person}{Yue Tan}, \bibinfo{person}{Guodong Long}, \bibinfo{person}{Lu Liu}, \bibinfo{person}{Tianyi Zhou}, \bibinfo{person}{Qinghua Lu}, \bibinfo{person}{Jing Jiang}, {and} \bibinfo{person}{Chengqi Zhang}.} \bibinfo{year}{2022}\natexlab{}.
\newblock \showarticletitle{Fedproto: Federated prototype learning across heterogeneous clients}. In \bibinfo{booktitle}{\emph{Proceedings of the AAAI Conference on Artificial Intelligence}}, Vol.~\bibinfo{volume}{36}. \bibinfo{pages}{8432--8440}.
\newblock


\bibitem[Venkatesan et~al\mbox{.}(2000)]%
        {venkatesan2000robust}
\bibfield{author}{\bibinfo{person}{Ramarathnam Venkatesan}, \bibinfo{person}{S-M Koon}, \bibinfo{person}{Mariusz~H Jakubowski}, {and} \bibinfo{person}{Pierre Moulin}.} \bibinfo{year}{2000}\natexlab{}.
\newblock \showarticletitle{Robust image hashing}. In \bibinfo{booktitle}{\emph{Proceedings 2000 International Conference on Image Processing (Cat. No. 00CH37101)}}, Vol.~\bibinfo{volume}{3}. IEEE, \bibinfo{pages}{664--666}.
\newblock


\bibitem[Wah et~al\mbox{.}(2011)]%
        {wah2011caltech}
\bibfield{author}{\bibinfo{person}{Catherine Wah}, \bibinfo{person}{Steve Branson}, \bibinfo{person}{Peter Welinder}, \bibinfo{person}{Pietro Perona}, {and} \bibinfo{person}{Serge Belongie}.} \bibinfo{year}{2011}\natexlab{}.
\newblock \showarticletitle{The caltech-ucsd birds-200-2011 dataset}.
\newblock  (\bibinfo{year}{2011}).
\newblock


\bibitem[Wang et~al\mbox{.}(2018)]%
        {wang2018additive}
\bibfield{author}{\bibinfo{person}{Feng Wang}, \bibinfo{person}{Jian Cheng}, \bibinfo{person}{Weiyang Liu}, {and} \bibinfo{person}{Haijun Liu}.} \bibinfo{year}{2018}\natexlab{}.
\newblock \showarticletitle{Additive margin softmax for face verification}.
\newblock \bibinfo{journal}{\emph{IEEE Signal Processing Letters}} \bibinfo{volume}{25}, \bibinfo{number}{7} (\bibinfo{year}{2018}), \bibinfo{pages}{926--930}.
\newblock


\bibitem[Wang et~al\mbox{.}(2017)]%
        {wang2017deep}
\bibfield{author}{\bibinfo{person}{Xiaofang Wang}, \bibinfo{person}{Yi Shi}, {and} \bibinfo{person}{Kris~M Kitani}.} \bibinfo{year}{2017}\natexlab{}.
\newblock \showarticletitle{Deep supervised hashing with triplet labels}. In \bibinfo{booktitle}{\emph{Computer Vision--ACCV 2016: 13th Asian Conference on Computer Vision, Taipei, Taiwan, November 20-24, 2016, Revised Selected Papers, Part I 13}}. Springer, \bibinfo{pages}{70--84}.
\newblock


\bibitem[Weiss et~al\mbox{.}(2008)]%
        {weiss2008spectral}
\bibfield{author}{\bibinfo{person}{Yair Weiss}, \bibinfo{person}{Antonio Torralba}, {and} \bibinfo{person}{Rob Fergus}.} \bibinfo{year}{2008}\natexlab{}.
\newblock \showarticletitle{Spectral hashing}.
\newblock \bibinfo{journal}{\emph{Advances in neural information processing systems}}  \bibinfo{volume}{21} (\bibinfo{year}{2008}).
\newblock


\bibitem[Wu et~al\mbox{.}(2024)]%
        {wu2024deepseek}
\bibfield{author}{\bibinfo{person}{Zhiyu Wu}, \bibinfo{person}{Xiaokang Chen}, \bibinfo{person}{Zizheng Pan}, \bibinfo{person}{Xingchao Liu}, \bibinfo{person}{Wen Liu}, \bibinfo{person}{Damai Dai}, \bibinfo{person}{Huazuo Gao}, \bibinfo{person}{Yiyang Ma}, \bibinfo{person}{Chengyue Wu}, \bibinfo{person}{Bingxuan Wang}, {et~al\mbox{.}}} \bibinfo{year}{2024}\natexlab{}.
\newblock \showarticletitle{Deepseek-vl2: Mixture-of-experts vision-language models for advanced multimodal understanding}.
\newblock \bibinfo{journal}{\emph{arXiv preprint arXiv:2412.10302}} (\bibinfo{year}{2024}).
\newblock


\bibitem[Xian et~al\mbox{.}(2018)]%
        {xian2018zero}
\bibfield{author}{\bibinfo{person}{Yongqin Xian}, \bibinfo{person}{Christoph~H Lampert}, \bibinfo{person}{Bernt Schiele}, {and} \bibinfo{person}{Zeynep Akata}.} \bibinfo{year}{2018}\natexlab{}.
\newblock \showarticletitle{Zero-shot learning—a comprehensive evaluation of the good, the bad and the ugly}.
\newblock \bibinfo{journal}{\emph{IEEE transactions on pattern analysis and machine intelligence}} \bibinfo{volume}{41}, \bibinfo{number}{9} (\bibinfo{year}{2018}), \bibinfo{pages}{2251--2265}.
\newblock


\bibitem[Xiao et~al\mbox{.}(2021)]%
        {xiao2021expert}
\bibfield{author}{\bibinfo{person}{Meng Xiao}, \bibinfo{person}{Ziyue Qiao}, \bibinfo{person}{Yanjie Fu}, \bibinfo{person}{Yi Du}, \bibinfo{person}{Pengyang Wang}, {and} \bibinfo{person}{Yuanchun Zhou}.} \bibinfo{year}{2021}\natexlab{}.
\newblock \showarticletitle{Expert knowledge-guided length-variant hierarchical label generation for proposal classification}. In \bibinfo{booktitle}{\emph{2021 ieee international conference on data mining (icdm)}}. IEEE, \bibinfo{pages}{757--766}.
\newblock


\bibitem[Xiao et~al\mbox{.}(2025)]%
        {xiao2025interdisciplinary}
\bibfield{author}{\bibinfo{person}{Meng Xiao}, \bibinfo{person}{Min Wu}, \bibinfo{person}{Ziyue Qiao}, \bibinfo{person}{Yanjie Fu}, \bibinfo{person}{Zhiyuan Ning}, \bibinfo{person}{Yi Du}, {and} \bibinfo{person}{Yuanchun Zhou}.} \bibinfo{year}{2025}\natexlab{}.
\newblock \showarticletitle{Interdisciplinary fairness in imbalanced research proposal topic inference: A hierarchical transformer-based method with selective interpolation}.
\newblock \bibinfo{journal}{\emph{ACM Transactions on Knowledge Discovery from Data}} \bibinfo{volume}{19}, \bibinfo{number}{2} (\bibinfo{year}{2025}), \bibinfo{pages}{1--21}.
\newblock


\bibitem[Xu et~al\mbox{.}(2020)]%
        {xu2020attribute}
\bibfield{author}{\bibinfo{person}{Wenjia Xu}, \bibinfo{person}{Yongqin Xian}, \bibinfo{person}{Jiuniu Wang}, \bibinfo{person}{Bernt Schiele}, {and} \bibinfo{person}{Zeynep Akata}.} \bibinfo{year}{2020}\natexlab{}.
\newblock \showarticletitle{Attribute prototype network for zero-shot learning}.
\newblock \bibinfo{journal}{\emph{Advances in Neural Information Processing Systems}}  \bibinfo{volume}{33} (\bibinfo{year}{2020}), \bibinfo{pages}{21969--21980}.
\newblock


\bibitem[Xu et~al\mbox{.}(2017)]%
        {xu2017attribute}
\bibfield{author}{\bibinfo{person}{Yahui Xu}, \bibinfo{person}{Yang Yang}, \bibinfo{person}{Fumin Shen}, \bibinfo{person}{Xing Xu}, \bibinfo{person}{Yuxuan Zhou}, {and} \bibinfo{person}{Heng~Tao Shen}.} \bibinfo{year}{2017}\natexlab{}.
\newblock \showarticletitle{Attribute hashing for zero-shot image retrieval}. In \bibinfo{booktitle}{\emph{2017 IEEE International Conference on Multimedia and Expo (ICME)}}. IEEE, \bibinfo{pages}{133--138}.
\newblock


\bibitem[Yang et~al\mbox{.}(2016)]%
        {yang2016zero}
\bibfield{author}{\bibinfo{person}{Yang Yang}, \bibinfo{person}{Yadan Luo}, \bibinfo{person}{Weilun Chen}, \bibinfo{person}{Fumin Shen}, \bibinfo{person}{Jie Shao}, {and} \bibinfo{person}{Heng~Tao Shen}.} \bibinfo{year}{2016}\natexlab{}.
\newblock \showarticletitle{Zero-shot hashing via transferring supervised knowledge}. In \bibinfo{booktitle}{\emph{Proceedings of the 24th ACM international conference on Multimedia}}. \bibinfo{pages}{1286--1295}.
\newblock


\bibitem[Yuan et~al\mbox{.}(2020)]%
        {yuan2020central}
\bibfield{author}{\bibinfo{person}{Li Yuan}, \bibinfo{person}{Tao Wang}, \bibinfo{person}{Xiaopeng Zhang}, \bibinfo{person}{Francis~EH Tay}, \bibinfo{person}{Zequn Jie}, \bibinfo{person}{Wei Liu}, {and} \bibinfo{person}{Jiashi Feng}.} \bibinfo{year}{2020}\natexlab{}.
\newblock \showarticletitle{Central similarity quantization for efficient image and video retrieval}. In \bibinfo{booktitle}{\emph{Proceedings of the IEEE/CVF conference on computer vision and pattern recognition}}. \bibinfo{pages}{3083--3092}.
\newblock


\bibitem[Zeng et~al\mbox{.}(2025)]%
        {zeng2025enhancing}
\bibfield{author}{\bibinfo{person}{Hansheng Zeng}, \bibinfo{person}{Yuqi Li}, \bibinfo{person}{Ruize Niu}, \bibinfo{person}{Chuanguang Yang}, {and} \bibinfo{person}{Shiping Wen}.} \bibinfo{year}{2025}\natexlab{}.
\newblock \showarticletitle{Enhancing spatiotemporal prediction through the integration of Mamba state space models and Diffusion Transformers}.
\newblock \bibinfo{journal}{\emph{Knowledge-Based Systems}} (\bibinfo{year}{2025}).
\newblock


\bibitem[Zhang et~al\mbox{.}(2019)]%
        {zhang2019zero}
\bibfield{author}{\bibinfo{person}{Haofeng Zhang}, \bibinfo{person}{Yang Long}, {and} \bibinfo{person}{Ling Shao}.} \bibinfo{year}{2019}\natexlab{}.
\newblock \showarticletitle{Zero-shot hashing with orthogonal projection for image retrieval}.
\newblock \bibinfo{journal}{\emph{Pattern Recognition Letters}}  \bibinfo{volume}{117} (\bibinfo{year}{2019}), \bibinfo{pages}{201--209}.
\newblock


\bibitem[Zheng and Gupta(2022)]%
        {zheng2022semantic}
\bibfield{author}{\bibinfo{person}{Shen Zheng} {and} \bibinfo{person}{Gaurav Gupta}.} \bibinfo{year}{2022}\natexlab{}.
\newblock \showarticletitle{Semantic-guided zero-shot learning for low-light image/video enhancement}. In \bibinfo{booktitle}{\emph{Proceedings of the IEEE/CVF Winter conference on applications of computer vision}}. \bibinfo{pages}{581--590}.
\newblock


\bibitem[Zhou et~al\mbox{.}(2019)]%
        {zhou2019angular}
\bibfield{author}{\bibinfo{person}{Chang Zhou}, \bibinfo{person}{Lai-Man Po}, \bibinfo{person}{Wilson~YF Yuen}, \bibinfo{person}{Kwok~Wai Cheung}, \bibinfo{person}{Xuyuan Xu}, \bibinfo{person}{Kin~Wai Lau}, \bibinfo{person}{Yuzhi Zhao}, \bibinfo{person}{Mengyang Liu}, {and} \bibinfo{person}{Peter~HW Wong}.} \bibinfo{year}{2019}\natexlab{}.
\newblock \showarticletitle{Angular deep supervised hashing for image retrieval}.
\newblock \bibinfo{journal}{\emph{IEEE Access}}  \bibinfo{volume}{7} (\bibinfo{year}{2019}), \bibinfo{pages}{127521--127532}.
\newblock


\bibitem[Zhu et~al\mbox{.}(2024b)]%
        {zhu2024distribution}
\bibfield{author}{\bibinfo{person}{Haowei Zhu}, \bibinfo{person}{Ling Yang}, \bibinfo{person}{Jun-Hai Yong}, \bibinfo{person}{Hongzhi Yin}, \bibinfo{person}{Jiawei Jiang}, \bibinfo{person}{Meng Xiao}, \bibinfo{person}{Wentao Zhang}, {and} \bibinfo{person}{Bin Wang}.} \bibinfo{year}{2024}\natexlab{b}.
\newblock \showarticletitle{Distribution-aware data expansion with diffusion models}.
\newblock \bibinfo{journal}{\emph{arXiv preprint arXiv:2403.06741}} (\bibinfo{year}{2024}).
\newblock


\bibitem[Zhu et~al\mbox{.}(2022)]%
        {zhu2022lower}
\bibfield{author}{\bibinfo{person}{Xiaosu Zhu}, \bibinfo{person}{Jingkuan Song}, \bibinfo{person}{Yu Lei}, \bibinfo{person}{Lianli Gao}, {and} \bibinfo{person}{Hengtao Shen}.} \bibinfo{year}{2022}\natexlab{}.
\newblock \showarticletitle{A lower bound of hash codes' performance}.
\newblock \bibinfo{journal}{\emph{Advances in Neural Information Processing Systems}}  \bibinfo{volume}{35} (\bibinfo{year}{2022}), \bibinfo{pages}{29166--29178}.
\newblock


\bibitem[Zhu et~al\mbox{.}(2024a)]%
        {zhu2024multivariate}
\bibfield{author}{\bibinfo{person}{Zhihong Zhu}, \bibinfo{person}{Xuxin Cheng}, \bibinfo{person}{Yunyan Zhang}, \bibinfo{person}{Zhaorun Chen}, \bibinfo{person}{Qingqing Long}, \bibinfo{person}{Hongxiang Li}, \bibinfo{person}{Zhiqi Huang}, \bibinfo{person}{Xian Wu}, {and} \bibinfo{person}{Yefeng Zheng}.} \bibinfo{year}{2024}\natexlab{a}.
\newblock \showarticletitle{Multivariate Cooperative Game for Image-Report Pairs: Hierarchical Semantic Alignment for Medical Report Generation}. In \bibinfo{booktitle}{\emph{International Conference on Medical Image Computing and Computer-Assisted Intervention}}. Springer, \bibinfo{pages}{303--313}.
\newblock


\end{thebibliography}

\end{document}